\documentclass[11pt]{article}

\usepackage[margin=1in]{geometry}
\usepackage[T1]{fontenc}
\usepackage[utf8]{inputenc}
\usepackage{lmodern}
\usepackage[hyphens]{url}
\usepackage{graphicx}
\usepackage{natbib}
\usepackage{caption}
\usepackage{booktabs}
\usepackage{amsfonts}
\usepackage{nicefrac}
\usepackage{microtype}
\usepackage{xcolor}
\usepackage{hyperref}
\hypersetup{
    colorlinks=true,
    citecolor=blue,
    linkcolor=red,
    urlcolor=blue
}
\usepackage{titling}
\thanksmarkseries{arabic}

\title{Cost-Aware Multi-Objective Bandits: Theory and Application to Budgeted LLM Configuration Evaluation}

\author{
  Bo Xue\thanks{City University of Hong Kong.
  Email: \texttt{boxue4-c@my.cityu.edu.hk}}
  \quad
  Zhi Hong\thanks{South China University of Technology.
  Email: \texttt{zhih1008@gmail.com}}
  \quad
  Jiayi Li\thanks{Zhejiang University.
  Email: \texttt{LJY@bupt.edu.cn}}
  \quad
  Yuanyu Wan\thanks{Zhejiang University.
  Email: \texttt{wanyy@zju.edu.hk}}
  \quad
  Ji Cheng\thanks{City University of Hong Kong.
  Email: \texttt{J.Cheng@my.cityu.edu.hk}}
  \quad
  Shuang Qiu{\color{blue}${}^\dag$}\thanks{City University of Hong Kong.
  Email: \texttt{shuanqiu@cityu.edu.hk}}
}

\usepackage[symbol]{footmisc}
\usepackage{amsmath}
\usepackage{paralist,amssymb,mathtools}
\usepackage{multirow}
\usepackage{bm}
\usepackage{algorithm,algorithmic}
\usepackage{color}

\newtheorem{thm}{Theorem}

\newtheorem{lem}{Lemma}
\newtheorem{cor}{Corollary}

\def \E {\mathrm{E}}
\def \Pr {\mathrm{Pr}}

\def \r {\bm{r}}
\def \bmu {\bm{\mu}}
\def \U {\mathbf{U}}

    \makeatletter
\def\@fnsymbol#1{\ensuremath{\ifcase#1\or *\or \dagger\or \ddagger\or
   \mathsection\or \mathparagraph\or \natural\or \sharp \or\dagger\dagger
   \or \ddagger\ddagger \else\@ctrerr\fi}}
    \makeatother
    
\begin{document}

\date{\today}

\footnotetext[2]{Corresponding Author}

\maketitle

\begin{abstract}
Large language model (LLM) configuration evaluation is challenging due to limited evaluation budgets, varying costs, and multiple competing objectives. In this paper, we formulate LLM configuration evaluation as a cost-aware multi-objective bandit problem, where each configuration evaluation incurs a configuration-dependent cost and yields a noisy vector-valued outcome. Under this framework, we study two fundamental problems: online configuration selection and Pareto configuration identification. For online configuration selection, we propose a hypervolume-based UCB algorithm that optimizes an optimistic hypervolume-per-cost index. We establish a budgeted regret bound of order $O\bigl(\sum_{i\ne i^\star}\frac{\log B}{\Delta_i}\bigr)$, where $B$ is the evaluation budget, $i^\star$ is the optimal configuration in terms of hypervolume efficiency, and $\Delta_i$ is the corresponding efficiency gap of configuration $i$. This bound retains the logarithmic budget dependence of classical single-objective budgeted bandits. For fixed-budget Pareto identification, we develop a cost-aware empirical gap elimination algorithm and prove that its error probability is of order $O\bigl(\exp(-\frac{B}{H_{\mu,c}})\bigr)$, where $H_{\mu,c}$ is a cost-aware Pareto identification complexity depending on configuration costs and Pareto classification gaps. This error probability decays exponentially with the evaluation budget and recovers the standard Pareto set identification guarantee when all configuration costs are identical. Experiments on LLM configuration evaluation tasks demonstrate that the proposed framework enables efficient online decision-making and accurate cost-aware Pareto identification under limited budgets.

\end{abstract}

\section{Introduction}
In real-world large language model (LLM) applications, users must select suitable configurations, such as base models, prompt templates, and retrieval settings, while operating under limited evaluation budgets. Evaluating a configuration may require model inference, long-form generation, LLM-as-a-judge scoring, or human assessment, leading to substantially different monetary costs across configurations \citep{Zheng:2023-LLM-Judge,Chang:2024-survey-eval,Li:2025-judge}. For example, larger models and chain-of-thought prompting may improve task performance but often consume more tokens \citep{Wei:2022-CoT,Shi:2024-decoding,Zhou:2024-efficient-inference}. Moreover, LLM evaluation is inherently multi-objective: beyond task accuracy, practitioners often care about latency and efficiency \citep{Liang:2023-HELM,Chang:2024-survey-eval}. Therefore, reliable LLM configuration evaluation requires adaptive allocation of a limited budget across candidate configurations while accounting for both varying evaluation costs and multiple quality objectives.

A straightforward approach to identifying the optimal LLM configuration is to exhaustively evaluate every candidate on the entire validation set and select the one with the highest empirical performance. This protocol aligns with standard benchmark-based evaluation practices, where models or systems are assessed across a predefined collection of tasks, examples, and metrics \citep{Liang:2023-HELM,Chang:2024-survey-eval}. Despite its simplicity and statistical reliability, exhaustive evaluation is increasingly impractical for modern LLM applications, as assessing a large number of configurations requires a substantial number of inference queries, leading to considerable monetary costs, token consumption, and computational overhead \citep{Zheng:2023-LLM-Judge,Li:2025-judge,Feng:2025-sample-evaluation}.

Recent studies have therefore explored adaptive evaluation strategies to improve evaluation efficiency. \citet{Zhou:2025} formulate language model evaluation as a bandit-based adaptive sampling problem. Similarly, \citet{Shi:2024} leverage best-arm identification techniques for prompt selection, with the goal of identifying high-performing prompts under a limited evaluation budget. These studies show that bandit-based adaptive sampling can significantly reduce evaluation costs by allocating more evaluation resources to promising candidates, instead of uniformly evaluating all candidates across all validation instances.

Despite these advances, existing adaptive evaluation methods still leave three limitations.
\textbf{(a)} They often overlook configuration-dependent evaluation costs, even though differences in model scale,  reasoning strategy, and retrieval pipeline can lead to substantial variation in token consumption and computational overhead \citep{Chen:2024-FrugalGPT,Ong:2025-RouteLLM}.
\textbf{(b)} Most existing formulations are restricted to single-objective optimization, focusing on a scalar performance metric or ranking criterion. 
However, practical LLM evaluation often requires balancing multiple objectives, including task effectiveness, robustness and latency \citep{Liang:2023-HELM,Chang:2024-survey-eval}. 
\textbf{(c)} Bandit-based LLM evaluation methods typically follow a pure-exploration formulation, assessing only the final configuration selected \citep{Polo:2024-PromptEval,Zhou:2025}. They therefore overlook the quality of intermediate decisions, which directly affects cumulative utility in continuous evaluation systems \citep{Lattimore:2020,Ong:2025-RouteLLM,Ashizawa:2025-Prompt}.

Motivated by these limitations, we formulate LLM configuration evaluation as a cost-aware multi-objective bandit problem with a finite evaluation budget. Each configuration is modeled as an arm, and each evaluation consumes a configuration-dependent cost while producing stochastic vector-valued feedback. This formulation captures two central characteristics of practical LLM evaluation: the need to allocate a limited budget across configurations with varying costs and the need to optimize multiple performance criteria. 

To the best of our knowledge, this is the first study of cost-aware multi-objective bandits. We consider two complementary learning objectives: online configuration selection through regret minimization and Pareto-optimal configuration identification under a fixed budget. Our main contributions are summarized as follows:
\begin{itemize}
\item We propose a hypervolume-based UCB algorithm for cost-aware online configuration selection, which optimizes a hypervolume-per-cost index to balance multi-objective performance and evaluation cost. Its regret is bounded by $O\bigl(\sum_{i\ne i^\star}\frac{\log B}{\Delta_i}\bigr)$, where $B$ is the evaluation budget, $i^\star$ is the optimal configuration in terms of hypervolume efficiency, and $\Delta_i$ is the corresponding efficiency gap of configuration $i$. This bound matches the single-objective result in both budget and gap dependence \citep{Ding:2013}.

\item We develop a cost-aware Pareto configuration identification algorithm, which allocates samples according to evaluation costs and progressively eliminates configurations using empirical Pareto gaps. Its error probability is bounded by $O\bigl(\exp(-\frac{B}{H_{\mu,c}})\bigr)$, where $H_{\mu,c}$ is a cost-aware Pareto identification complexity depending on configuration costs and Pareto classification gaps. This error probability decays exponentially with the evaluation budget and recovers the standard error rate for fixed-budget Pareto set identification when all configurations have identical evaluation costs \citep{Kone:2024}.

\item We conduct experiments on LLM configuration evaluation tasks involving different models, prompts and decoding parameters. Comparisons with cost-insensitive bandit algorithms and single-objective selection strategies, our methods demonstrate the benefits of jointly accounting for evaluation cost and multiple performance objectives.
\end{itemize}

\section{Related Work}
This section reviews the literature on budgeted bandits, multi-objective bandits, and recent bandit-based methods for LLMs.


\paragraph{Budgeted Bandits.}
 Classical bandit research has primarily considered two learning objectives: regret minimization, which evaluates the quality of decisions made throughout the learning process \citep{Lai:1985,Auer:2002}, and best-arm identification, which focuses on the quality of the final recommendation after exploration \citep{Audibert:2010,Gabillon:2012,Jamieson:2014}. \citet{Ding:2013} formalized the budgeted multi-armed bandit with stochastic costs, which captures the reward-cost trade-off through an expected reward-per-cost criterion and provides logarithmic regret guarantees. Subsequent works have extended this line in several directions. \citet{Xia:2015-BMAB-TS} developed Thompson sampling algorithms for budgeted bandits with random costs and showed that Bayesian sampling achieves logarithmic regret in the budget. \citet{Xia:2016-MPBMAB} investigated the multiple-play setting, in which several arms are selected in each round under a shared budget. \citet{Li:2017-InfiniteBMAB} considered budgeted bandits with infinitely many arms, highlighting the additional exploration cost incurred when searching over a large action space. A broader related framework is bandits with knapsacks, which generalizes budgeted bandits to multiple resource constraints \citep{Badanidiyuru:2018-BwK,Badanidiyuru:2014-ResourcefulCB,Immorlica:2022-AdversarialBwK,Sankararaman:2018-CombinatorialBwK,Liu:2022-NonstationaryBwK}. These studies provide a principled foundation for decision-making under resource constraints, but they mainly focus on scalar rewards. 

\paragraph{Multi-Objective Bandits.}
Multi-objective bandits address sequential decision-making with vector-valued rewards. Existing studies have considered several performance criteria, including scalarization-based regret \citep{Drugan:2013}, Pareto regret \citep{Lu:2019}, hypervolume regret \citep{Zhang:2024-HVScalarization}, and Pareto set identification (PSI) \citep{Auer:2016}. A relevant line optimizes the hypervolume indicator. 
\citet{Zhang:2020-RHS} introduced random hypervolume scalarizations for multi-objective black-box optimization, and \citet{Zhang:2024-HVScalarization} later established optimal sublinear hypervolume regret guarantees for this approach. Another closely related line investigates PSI, whose goal is to identify all non-dominated arms. \citet{Auer:2016} studied PSI under stochastic bandit feedback, \citet{Cyrille:2023} developed adaptive algorithms for relaxed fixed-confidence PSI, and \citet{Kone:2024} proposed successive-rejects-based algorithms for the fixed-budget setting. More recent work has also examined the sequential learning of Pareto fronts in multi-objective bandits \citep{Crepon:2024}. Furthermore, multi-objective reinforcement learning with bandit feedback \citep{qiu2024traversing}  extends the bandit problem to Markov decision processes. Although these studies provide important foundations for multi-objective exploration, they typically constrain the total number of sampling rounds without explicitly accounting for arm-dependent evaluation costs.

\paragraph{Bandits for LLMs.}
Bandit learning has recently been explored in several LLM-related problems, including prompt optimization, model selection, and budget-efficient evaluation. ProTeGi incorporates a bandit-based selection mechanism into beam search to identify promising prompt candidates \citep{Pryzant:2023-ProTeGi}. \citet{Lin:2024-INSTINCT} employ neural bandits with transformer-based representations for instruction optimization, whereas \citet{Wu:2024-EASE} apply neural bandits to order-aware in-context exemplar selection. More recently, \citet{Hong:2026-MASPOB} propose MASPOB for prompt optimization in LLM-based multi-agent systems, where UCB-style exploration is combined with graph neural networks to capture topology-induced coupling among agents. Another related line formulates model, prompt, or evaluation selection as a bandit problem. \citet{Shi:2024} connect prompt selection with fixed-budget best arm identification, while \citet{Poon:2025-online-llm-selection} study online LLM selection under contextual information and cost-quality trade-offs. For budget-efficient evaluation, \citet{Zhou:2025} integrate multi-armed bandits with low-rank matrix factorization to adaptively allocate evaluations and identify high-performing configurations using fewer observations. These studies demonstrate the potential of bandit methods to reduce the costs of prompt search, model selection, and performance evaluation. Nevertheless, most existing approaches either optimize a single scalar objective or focus primarily on offline best-arm identification.

\section{Preliminaries}
We consider LLM configuration evaluation with multiple competing objectives under a total evaluation budget $B>0$.

At each round $t=1,2,\ldots$, the learner selects a configuration $i_t\in[K]\coloneqq\{1,\ldots,K\}$. The selected configuration is evaluated on a validation sample, such as a question drawn from GSM8K \citep{GSM8K:2021} or PIQA \citep{PIQA:2020}, and produces a reward vector
$$
\r_{i_t,t}=\bigl(r_{i_t,t}^{(1)},\ldots,r_{i_t,t}^{(D)}\bigr)\in[0,1]^D,
$$
where each coordinate corresponds to one normalized objective, such as task accuracy, inference efficiency, or other deployment-relevant criteria. All objectives are formulated so that larger values are preferred. For each configuration $i\in[K]$, its reward vectors are i.i.d. with unknown mean
$$
\bmu_i=\bigl(\mu_i^{(1)},\ldots,\mu_i^{(D)}\bigr)=\E[\r_{i,t}].
$$

In addition to producing a multi-objective reward, each evaluation consumes part of the available budget. This cost may represent monetary API expenditure, token consumption, or human-annotation effort. Although these criteria can also be incorporated into the reward vector, the evaluation cost determines the total number of evaluations. 

We distinguish between two cost models for the two learning settings. In online cost-aware configuration selection, the evaluation cost may be random. Specifically, evaluating configuration $i$ at round $t$ incurs a cost
$$
c_{i,t}\in[\lambda,1], \quad \mu_i^c=\E[c_{i,t}], \quad 0<\lambda\le 1.
$$

For fixed-budget Pareto set identification, we instead consider deterministic and known evaluation costs,
$$
c_i\in[\lambda,1].
$$
This model is appropriate when the cost of a single evaluation can be estimated in advance, for example, a fixed inference protocol, or a standardized human-evaluation procedure.

\subsection{Budgeted Online Configuration Selection}
Let $C_t=\sum_{s=1}^{t} c_{i_s,s}$ denote the cumulative cost after $t$ evaluations. Given an evaluation budget $B>0$, the number of evaluations completed within the budget is
$$
    T(B)
    =
    \max\{t\ge 0:C_t\le B\}.
$$

We quantify the multi-objective performance of each configuration using the hypervolume indicator. With the zero vector as the reference point, the single-point hypervolume of configuration $i$ is
$$
    H_i
    =
    H(\bmu_i)
    =
    \operatorname{HV}(\bmu_i)
    =
    \prod_{j=1}^{D}\mu_i^{(j)}.
$$


To account for varying evaluation costs, we define the hypervolume efficiency for any $i\in[K]$ as $\nu_i=\frac{H_i}{\mu_i^c}$. Let $i^\star=\arg\max_{i\in[K]}\nu_i$ denote the optimal configuration and define the corresponding efficiency gap as $\Delta_{i} =\nu_{i^\star}-\nu_i$. The hypervolume-efficiency regret under budget $B$ is
$$
    R_B^{\operatorname{HV}}
    =
    \sum_{t=1}^{T(B)}\Delta_{i_t}.
$$
We further consider the standard hypervolume regret
$$
    R_B^{\operatorname{std}}
    =
    R_B^\star
    -\sum_{t=1}^{T(B)}H_{i_t},
$$
where $R_B^\star$ denotes the maximum expected cumulative hypervolume attainable under budget $B$. 

Both regret notions measure cumulative performance under the budget constraint, while $R_B^{\operatorname{HV}}$ further emphasizes hypervolume gain per unit cost.

\subsection{Budgeted Pareto Configuration Identification}
We also consider the fixed-budget identification setting, where the learner aims to recover the Pareto-optimal configuration set under a prescribed budget $B>0$.

Following \citet{Kone:2024}, for any $\bm u,\bm v\in\mathbb R^D$, we write $\bm u\preceq\bm v$ if $u^{(j)}\le v^{(j)}$ for any $j\in[D]$ and $\bm u\prec\bm v$ if $u^{(j)}< v^{(j)}$ for any $j\in[D]$. Configuration $j$ strictly Pareto dominates configuration $i$ if $\bmu_i\prec\bmu_j$. The true Pareto set is therefore
$$
    \mathcal P^\star
    =
    \left\{
    i\in[K]:
    \nexists j\in[K]\ \text{such that}\ 
    \bmu_i\prec\bmu_j
    \right\}.
$$

The learner adaptively evaluates configurations with deterministic costs $\{c_i\}_{i=1}^K$ and, upon exhausting the available budget, outputs an estimate $\widehat{\mathcal P}_B$. Its performance is measured by the misidentification probability
$$
    \Pr
    \left(
        \widehat{\mathcal P}_B
        \ne
        \mathcal P^\star
    \right).
$$

To characterize the difficulty of Pareto classification, for any pair $i,j\in[K]$, define
\begin{equation*}
\begin{aligned}
m(i,j)
&=
\min_{d\in[D]}
\left(
    \mu_j^{(d)}
    -
    \mu_i^{(d)}
\right),\\
M(i,j)
&=
\max_{d\in[D]}
\left(
    \mu_i^{(d)}
    -
    \mu_j^{(d)}
\right).
\end{aligned}
\end{equation*}
Following the gap construction of Empirical Gap Elimination
\citep{Kone:2024}, define
\[
    \Delta_i^-
    =
    \max_{j\in[K]\setminus\{i\}}m(i,j)
\]
and, for a Pareto-optimal configuration $i\in\mathcal P^\star$,
\[
\begin{aligned}
    \Delta_i^+
    &=
    \min_{j\in[K]\setminus\{i\}}
    \left[
        M(i,j)
        \wedge
        \left(
            M(j,i)^+
            +
            \left(\Delta_j^-\right)^+
        \right)
    \right],
\end{aligned}
\]
where $x^+=\max\{x,0\}$ and $a\wedge b=\min\{a,b\}$. Thus, the classification gap is
\[
    \gamma_i
    =
    \begin{cases}
    \Delta_i^-,
    & i\notin\mathcal P^\star,\\[1ex]
    \Delta_i^+,
    & i\in\mathcal P^\star.
    \end{cases}
\]
For a dominated configuration, $\gamma_i$ measures the strongest uniform domination margin.  For a Pareto-optimal configuration, the two terms in $\Delta_i^+$ jointly account for its separation from other Pareto arms and from dominated competitors. 

We assume $\gamma_i>0$ for every $i$, as exact identification is not statistically well posed when a classification gap is zero. Let $\gamma_{(1)}\le\gamma_{(2)}\le\cdots\le\gamma_{(K)}$ denote the ordered classification gaps. For any $\mathcal A\subseteq[K]$, its aggregate cost is
$
  C(\mathcal A)
  =
  \sum_{i\in\mathcal A} c_i.
$
The deterministic-cost problem complexity is
\[
    H_{\boldsymbol\mu,c}
    =
    \max_{\emptyset\ne\mathcal A\subseteq[K]}
    \frac{
        C(\mathcal A)
    }{
        \gamma_{(|\mathcal A|)}^2
    }.
\]
This quantity jointly captures the statistical difficulty of distinguishing Pareto-optimal and dominated configurations and the cost of evaluating the configurations that remain unresolved.

\section{Algorithms}
In this section, we present two cost-aware algorithms for multi-objective bandits, addressing online selection and Pareto set identification, respectively.

\subsection{Cost-Aware Hypervolume UCB}
We first introduce \textsc{CoHV-UCB} for budgeted online multi-objective configuration selection. Following the optimism-in-the-face-of-uncertainty principle, the algorithm constructs an optimistic reward vector for each configuration and evaluates it using the hypervolume indicator. It then divides the optimistic hypervolume by a lower confidence bound on the expected evaluation cost, yielding an optimistic hypervolume-per-cost index for configuration selection.

\textsc{CoHV-UCB} starts with an initialization phase. It pulls each arm once, observes its reward vector and evaluation cost, and initializes the empirical reward and cost estimates. At the beginning of each round, \textsc{CoHV-UCB} first computes empirical estimates for every arm. For each objective $d\in[D]$, the empirical mean reward of arm $i$ is
$$
    \bar r_{i,t}^{(d)}
    =
    \frac{1}{n_{i,t}}
    \sum_{s<t:i_s=i} r_{i,s}^{(d)}, 
$$
where $n_{i,t} = \sum_{s<t}\mathbb I\{i_s=i\}$ is the number of times arm $i$ has been evaluated before round $t$. 

Given the budget $B$, define $T_B=\left\lceil \frac{B}{\lambda}\right\rceil+1.$ Because each evaluation incurs a cost of at least $\lambda$, $T_B$ provides a deterministic upper bound on the effective evaluation horizon. Accordingly, the confidence radius for arm $i$ at round $t$ is
$$
    \beta_{i,t}
    =
    \sqrt{
    \frac{\alpha\log T_B}{n_{i,t}}
    },
    \quad
    \alpha\ge 2 .
$$

Using this confidence radius, \textsc{CoHV-UCB} constructs an optimistic estimate of the $d$-th objective:
$$
    U_{i,t}^{(d)}
    =
    \min
    \left\{
    1,\,
    \bar r_{i,t}^{(d)}+\beta_{i,t}
    \right\},
    \quad
    d=1,\ldots,D.
$$
Clipping the estimate ensures that the resulting objective vector remains within the normalized reward domain $[0,1]^D$. The algorithm then estimates the evaluation cost of each arm. The empirical mean cost of arm $i$ is
$$
    \bar c_{i,t}
    =
    \frac{1}{n_{i,t}}
    \sum_{s<t:i_s=i} c_{i,s}.
$$
Because the selection index is inversely proportional to the estimated cost, underestimating the denominator can excessively inflate the index. To avoid this issue, \textsc{CoHV-UCB} employs the truncated lower confidence bound
$$
    \underline c_{i,t}
    =
    \max
    \left\{
    \frac{\lambda}{2},\,
    \bar c_{i,t}-\beta_{i,t}
    \right\}.
$$

\begin{algorithm}[t]
\caption{\textsc{CoHV-UCB}: Cost-aware HyperVolume UCB}
\label{alg:cohv_ucb}
\begin{algorithmic}[1]
\REQUIRE Budget $B$, number of arms $K$, cost lower bound $\lambda$
\STATE Pull each arm $i\in[K]$ once and set $t=K$
\WHILE{$\sum_{s=1}^{t} c_{a_s,s} \le B$}
    \STATE Set $t=t+1$.
    \FOR{each arm $i\in[K]$}
        \STATE Compute the optimistic vector $\U_{i,t}$ with entries
        \[
        U_{i,t}^{(d)}
        =
        \min\left\{
        1,\,
        \bar r_{i,t}^{(d)}+\beta_{i,t}
        \right\},
        \quad d=1,\ldots,D
        \]
        \STATE Compute the lower confidence bound of the evaluation cost: 
        $\underline c_{i,t}
        =
        \max\left\{
        \lambda/2,\,
        \bar c_{i,t}-\beta_{i,t}
        \right\}$
        \STATE Compute the cost-aware hypervolume index:
        \[
        I_{i,t}
        =
        \frac{
        \operatorname{HV}(\U_{i,t})
        }{
        \underline c_{i,t}
        }
        \]
    \ENDFOR
    \STATE Pull the arm with the largest index:
    \[
    i_t
    =
    \arg\max_{i\in[K]}I_{i,t}
    \]
\ENDWHILE
\end{algorithmic}
\end{algorithm}

After constructing the optimistic rewards and the lower confidence estimate of the evaluation cost, \textsc{CoHV-UCB} computes the cost-aware hypervolume UCB index,
$
    I_{i,t}
    =
    \frac{
    \operatorname{HV}(\mathbf U_{i,t})
    }{
    \underline c_{i,t}
    } .
$
The numerator measures the optimistic multi-objective quality of the configuration, while the denominator measures a conservative estimate of its evaluation cost. Thus, $I_{i,t}$ provides an optimistic estimate of the hypervolume efficiency.

At round $t$, \textsc{CoHV-UCB} selects the configuration with the largest index,
$
    i_t
    =
    \arg\max_{i\in[K]} I_{i,t}.
$
It then evaluates configuration $i_t$, observes the corresponding reward vector and evaluation cost, and updates the empirical reward and cost. This procedure continues until the budget $B$ is exhausted.

We next establish the regret guarantee for \textsc{CoHV-UCB}. The resulting regret retains the logarithmic budget dependence and inverse-gap dependence of cost-aware single-objective budgeted bandits \citep{Ding:2013}, up to additional factors arising from the hypervolume objective.

\begin{thm}[Regret bound of \textsc{CoHV-UCB}]
\label{thm:cohv_ucb_regret}
Let $\alpha\ge 2$ and define
$
    C_{\lambda,H}
    =
    \frac{4D}{\lambda}
    +
    \frac{4}{\lambda^2}.
$
For each suboptimal arm $i\ne i^\star$, let $n_i(B)$ denote the number of times arm $i$ is pulled before the budget $B$ is exhausted. Then, under \textsc{CoHV-UCB},
$$
\begin{aligned}
    \mathbb E[n_i(B)]
    \le
    1
    +
    \left\lceil
    \frac{
        4\alpha C_{\lambda,H}^2\log T_B
    }{
        \Delta_i^2
    }
    \right\rceil
    +
    2K(D+1)T_B^{2-2\alpha}.
\end{aligned}
$$
Consequently, the hypervolume-efficiency regret satisfies
$$
\begin{aligned}
    \mathbb E[R_B^{\operatorname{HV}}]
    =
    \sum_{i\ne i^\star}
    \Delta_i\,\mathbb E[n_i(B)]
    =
    O\left(
    \sum_{i\ne i^\star}
    \frac{
        C_{\lambda,H}^2\log B
    }{
        \Delta_i
    }
    \right).
\end{aligned}
$$
\end{thm}

Theorem~\ref{thm:cohv_ucb_regret} bounds the online loss in hypervolume per unit cost. We next translate this result into a standard budgeted hypervolume regret bound, which compares the cumulative hypervolume with that of the optimal budgeted policy.

\begin{cor}[Standard regret bound of \textsc{CoHV-UCB}]
\label{cor:standard_cohv_regret}
Under the conditions of Theorem~\ref{thm:cohv_ucb_regret}, the standard budgeted hypervolume regret of \textsc{CoHV-UCB} satisfies
$$
    \mathbb E[R_B^{\operatorname{std}}]
    =
    O\left(
    \sum_{i\ne i^\star}
    \frac{
        C_{\lambda,H}^2\log B
    }{
        \Delta_i
    }
    \right).
$$
\end{cor}



\newcommand{\hbmu}{\widehat{\boldsymbol{\mu}}}
\newcommand{\calA}{\mathcal A}
\newcommand{\calC}{\mathcal C}
\newcommand{\calP}{\mathcal P}
\newcommand{\calR}{\mathcal R}
\newcommand{\calU}{\mathcal U}
\newcommand{\bbP}{\Pr}

\subsection{Cost-Aware Pareto Set Identification}
We now propose \textsc{CoPSI}, a cost-aware algorithm that allocates samples according to configuration-dependent costs and progressively identifies the Pareto-optimal configurations.

Throughout the procedure, \textsc{CoPSI} maintains three sets. At phase $r$, the active set $\mathcal A_r$ contains the configurations whose Pareto status has not yet been resolved. The set $\widehat{\mathcal P}$ collects configurations identified as Pareto-optimal, whereas $\widehat{\mathcal R}$ contains those identified as dominated. Initially, all configurations are active, so that $\mathcal A_1=[K]$, while $\widehat{\mathcal P}$ and $\widehat{\mathcal R}$ are both empty.

At the beginning of phase $r$, \textsc{CoPSI} computes the total cost of evaluating the active configurations,
$
    C_r
    =
    C(\mathcal A_r)
    =
    \sum_{i\in\mathcal A_r} c_i .
$
It then sets the cumulative sampling target to
$
    n_r
    =
    \left\lfloor
    \frac{B}{L_{K,\lambda}C_r}
    \right\rfloor ,
$
where
$
    L_{K,\lambda}
    =
    1+\sum_{\ell=2}^{K}\frac{1}{1+(\ell-1)\lambda}.
$
This allocation assigns the number of evaluations to each active configuration while accounting for the aggregate cost of evaluating the current active set. As the active set shrinks, the budget can support more samples for the remaining configurations.

Given the target $n_r$, \textsc{CoPSI} then brings every active configuration up to this sampling level. Specifically, each configuration $i\in\mathcal A_r$ is sampled $n_r-N_i$ additional times, where $N_i$ denotes the number of previous evaluations of configuration $i$. The empirical mean vector is then updated by
$$
    \widehat{\boldsymbol\mu}_{i,r}
    =
    \frac{1}{N_i}
    \sum_{s=1}^{N_i}\r_{i,s}.
$$

Using these empirical means, \textsc{CoPSI} constructs the empirical Pareto set over the active configurations:
$$
    \widehat{\mathcal P}_r^{\,\mathrm{emp}}
    =
    \left\{
    i\in\mathcal A_r:
    \nexists j\in \mathcal A_r\setminus\{i\}
    \text{ such that }
    \widehat{\boldsymbol\mu}_{i,r}
    \prec
    \widehat{\boldsymbol\mu}_{j,r}
    \right\}.
$$
Thus, every configuration outside $\widehat{\mathcal P}_r^{\mathrm{emp}}$ is empirically dominated by at least one active configuration, whereas every configuration in $\widehat{\mathcal P}_r^{\mathrm{emp}}$ is empirically nondominated.

\begin{algorithm}[t]
\caption{\textsc{CoPSI}: Cost-aware Pareto Set Identification}
\label{alg:copsi}
\begin{algorithmic}[1]
\REQUIRE Budget $B$, arm costs $\{c_i\}_{i=1}^K$
\STATE Initialize the active set $\mathcal A_1=[K]$, the accepted Pareto set $\widehat{\mathcal P}=\emptyset$, and the rejected set $\widehat{\mathcal R}=\emptyset$
\STATE Set $N_i=0$ for all $i\in[K]$
\STATE Set $L_{K,\lambda}=1+\sum_{\ell=2}^{K}\frac{1}{1+(\ell-1)\lambda}$
\STATE Set $n_0=0$

\FOR{$r=1,\ldots,K-1$}
    \STATE Set $k_r=|\mathcal A_r|$ and $C_r=C(\mathcal A_r)=\sum_{i\in\mathcal A_r}c_i$.
    
    \STATE Set the cumulative sampling target
    $
        n_r
        =
        \left\lfloor
        \frac{B}{L_{K,\lambda}C_r}
        \right\rfloor .
    $

    \FOR{$i\in\mathcal A_r$}
        \STATE Pull arm $i$ exactly $n_r-N_i$ additional times
        \STATE Update $N_i\leftarrow n_r$
        \STATE Compute the empirical mean
        $
            \widehat{\boldsymbol\mu}_{i,r}
            =
            \frac{1}{N_i}
            \sum_{s=1}^{N_i}\r_{i,s}.
        $
    \ENDFOR

    \STATE Compute the empirical Pareto set:
    $
        \widehat{\mathcal P}_r^{\mathrm{emp}}
        =
        \{
        i\in\mathcal A_r:
        \nexists j\in \mathcal A_r\setminus\{i\}
        \text{ such that }
        \widehat{\boldsymbol\mu}_{i,r}
        \prec
        \widehat{\boldsymbol\mu}_{j,r}
        \}.
    $

    \STATE Compute the empirical gap $\widehat\gamma_{i,r}$ for all $i\in\mathcal A_r$ by \eqref{eq:emp_classification_gap}

    \STATE Choose $e_r\in\arg\max_{i\in\mathcal A_r}\widehat\gamma_{i,r}$; if there are multiple maximizers, select one in $\mathcal A_r\setminus\widehat{\mathcal P}_r^{\mathrm{emp}}$ whenever possible

    \IF{$e_r\in\widehat{\mathcal P}_r^{\mathrm{emp}}$}
        \STATE Accept $e_r$:
        $
            \widehat{\mathcal P}
            \leftarrow
            \widehat{\mathcal P}\cup\{e_r\}.
        $
    \ELSE
        \STATE Reject $e_r$:
        $
            \widehat{\mathcal R}
            \leftarrow
            \widehat{\mathcal R}\cup\{e_r\}.
        $
    \ENDIF

    \STATE Update the active set
    $
        \mathcal A_{r+1}
        =
        \mathcal A_r\setminus\{e_r\}.
    $
\ENDFOR

\RETURN
$
    \widehat{\mathcal P}_B
    =
    \widehat{\mathcal P}
    \cup
    \mathcal A_K .
$
\end{algorithmic}
\end{algorithm}

After identifying the empirical Pareto set, \textsc{CoPSI} further quantifies how confidently each active configuration can be classified. For any phase $r$ and any two configurations $i,j\in\mathcal A_r$, define the empirical pairwise margins
\begin{equation*}
\begin{aligned}
    \widehat m_r(i,j)
    &=
    \min_{d\in[D]}
    \left(
        \widehat\mu_{j,r}^{(d)}
        -
        \widehat\mu_{i,r}^{(d)}
    \right),\\
    \widehat M_r(i,j)
    &=
    \max_{d\in[D]}
    \left(
        \widehat\mu_{i,r}^{(d)}
        -
        \widehat\mu_{j,r}^{(d)}
    \right).
\end{aligned}
\end{equation*}
Here, $\widehat m_r(i,j)$ measures the minimum coordinate-wise advantage of configuration $j$ over configuration $i$, while $\widehat M_r(i,j)$ measures the largest coordinate-wise advantage.

Let $x^+=\max\{x,0\}$. For each active configuration $i\in\mathcal A_r$, define the empirical dominatedness gap
\begin{equation*}
    \widehat\Delta_{i,r}^{-}
    =
    \max_{j\in\mathcal A_r\setminus\{i\}}
    \widehat m_r(i,j),
\end{equation*}
which measures the strongest empirical evidence that $i$ is dominated by another active configuration. We also define the empirical nondominatedness gap
\begin{equation*}
    \widehat\Delta_{i,r}^{+}
    =
    \min_{j\in\mathcal A_r\setminus\{i\}}
    \left[
        \widehat M_r(i,j)
        \wedge
        \left(
            \widehat M_r(j,i)^+
            +
            \left(\widehat\Delta_{j,r}^{-}\right)^+
        \right)
    \right].
\end{equation*}
This quantity evaluates the empirical evidence that $i$ should remain Pareto optimal, while also accounting for whether its competitors are themselves empirically dominated.

The empirical classification gap is then defined according to the empirical status of configuration $i$:
\begin{equation}\label{eq:emp_classification_gap}
    \widehat\gamma_{i,r}
    =
    \begin{cases}
    \widehat\Delta_{i,r}^{-},
    &
    i\notin\widehat{\mathcal P}_r^{\mathrm{emp}},
    \\[1ex]
    \widehat\Delta_{i,r}^{+},
    &
    i\in\widehat{\mathcal P}_r^{\mathrm{emp}}.
    \end{cases}
\end{equation}
Thus, $\widehat\gamma_{i,r}$ measures how reliably the current empirical evidence supports the classification of configuration $i$ as either dominated or nondominated. A larger value indicates that the status of the configuration is easier to determine, and hence the configuration can be safely removed from the active set.

Finally, \textsc{CoPSI} removes the active configuration with the largest empirical classification gap:
$
    e_r
    \in
    \arg\max_{i\in\mathcal A_r}
    \widehat\gamma_{i,r}.
$
If the maximizer is not unique, \textsc{CoPSI} selects one from 
$\mathcal A_r\setminus\widehat{\mathcal P}_r^{\mathrm{emp}}$ whenever possible. The selected configuration is accepted if it belongs to $\widehat{\mathcal P}_r^{\mathrm{emp}}$ and rejected otherwise.

If $e_r\in\widehat{\mathcal P}_r^{\,\mathrm{emp}}$, it is accepted into $\widehat{\mathcal P}$. Otherwise, it is rejected and added into $\widehat{\mathcal R}$. The classified configuration is then removed from the active set:
$$
    \mathcal A_{r+1}
    =
    \mathcal A_r\setminus\{e_r\}.
$$
The procedure runs for $K-1$ phases, leaving one unclassified arm in $\mathcal A_K$. This final arm is included in the output, so that
$\widehat{\mathcal P}_B=\widehat{\mathcal P}\cup\mathcal A_K$.

This elimination procedure is cost-aware because the sampling target in each phase depends on the total cost $C_r$ of the active configurations. It is also Pareto-aware because configurations are accepted or rejected according to empirical dominance relations. These properties make \textsc{CoPSI} particularly suitable for fixed-budget LLM configuration evaluation, where the objective is to identify a diverse set of nondominated configurations that captures different trade-offs among accuracy, latency, and efficiency.

We next establish the theoretical guarantee of \textsc{CoPSI}, showing that the probability of incorrectly identifying the Pareto set decreases exponentially with the available evaluation budget.

\begin{thm}[Fixed-budget error probability of \textsc{CoPSI}]
\label{thm:copsi} 
Let $K\ge 2$ and define
$
    L_{K,\lambda}
    =
    1+\sum_{\ell=2}^{K}
    \frac{1}{1+(\ell-1)\lambda}.
$
If the budget satisfies
$
    B
    \ge
    2L_{K,\lambda}H_{\boldsymbol\mu,c},
$
then the Pareto set estimate returned by \textsc{CoPSI} satisfies
$$
    \Pr
    \left(
    \widehat{\mathcal P}_B
    \ne
    \mathcal P^\star
    \right)
    \le
    2K^2D
    \exp
    \left(
    -
    \frac{
        B
    }{
        256L_{K,\lambda}H_{\boldsymbol\mu,c}
    }
    \right).
$$
\end{thm}

Theorem~\ref{thm:copsi} quantifies the difficulty of fixed-budget Pareto configuration identification through the cost-aware complexity measure $H_{\boldsymbol\mu,c}$. In particular, the probability of incorrectly identifying the Pareto set decreases exponentially with the available budget $B$. Moreover, when all configurations have unit evaluation costs, the bound recovers the standard exponential error rate for fixed-budget Pareto set identification \citep{Kone:2024}. 


\section{Experiments}
In this section, we conduct numerical experiments to evaluate the statistical performance and budget efficiency of the proposed methods on real LLM configuration data. 


\subsection{Experimental Setup}

\begin{table}[h]
\centering
\caption{Configuration-level dataset statistics.}
\small
\setlength{\tabcolsep}{3.5pt}
\begin{tabular}{lccccc}
\toprule
Data & $K$ & Acc. range & Latency (s) & Tokens & $|\mathcal P^\star|$\\
\midrule
GSM8K &108& .014--.165 & .432--8.854 &122.1--146.4&7\\
PIQA  &132& .462--.571 & .443--3.049 & 93.5--149.9&7\\
\bottomrule
\end{tabular}
\label{tab:data}
\end{table}

\paragraph{Datasets.} We use GSM8K \citep{GSM8K:2021} and PIQA \citep{PIQA:2020}, with 1,000 validation instances evaluated for every configuration. Each configuration combines a model, temperature, generation limit, and prompting strategy. GSM8K contains nine models and 108 configurations, while PIQA contains 11 models and 132 configurations. Table~\ref{tab:data} summarizes their configuration-level statistics.

\begin{figure}[h]
\centering
\setlength{\abovecaptionskip}{0cm}
\setlength{\belowcaptionskip}{0cm}
\includegraphics[width=8cm]{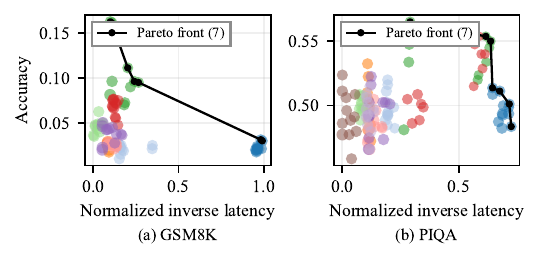}
\caption{Configuration landscapes on GSM8K and PIQA.}
\label{fig:landscape}
\end{figure}

\paragraph{Configuration landscapes.}
Beyond the aggregate ranges in Table~\ref{tab:data}, Figure~\ref{fig:landscape} shows the accuracy--efficiency trade-offs of individual configurations. Both axes are maximization objectives: efficiency is measured by normalized inverse latency. Point size represents mean token consumption, and the black curve marks the empirical Pareto front.

\paragraph{Evaluation settings.}
Online selection uses every configuration in each dataset. For exact Pareto identification, the complete sets contain many tied or near-tied empirical means; we therefore use a pre-specified 14-arm subset comprising seven common models at temperatures 0.5 and 1.0. It contains two Pareto arms on GSM8K and three on PIQA. Full-set approximate recovery and all preprocessing, replay, budget, and hyperparameter details are provided in the Appendix~\ref{app:additional-experiments}.

\paragraph{Baselines.}
For online selection, we compare \textsc{CoHV-UCB} with three baselines. \textsc{HV-UCB} removes the cost denominator from the proposed index and isolates the effect of cost awareness. \textsc{Accuracy-Cost-UCB} optimizes optimistic accuracy per token and isolates the value of modeling multiple objectives. Uniform round-robin allocation provides a non-adaptive reference. All three UCB-based methods use the same experimental confidence-radius calibration.

For fixed-budget identification, we compare \textsc{CoPSI} with PSI-SR, which ignores heterogeneous evaluation costs, and uniform allocation followed by empirical Pareto estimation. These comparisons separate the effect of cost-aware allocation from that of successive elimination and provide a strong non-adaptive reference when the arm gaps are relatively balanced.

\begin{figure*}[tb]
\centering
\setlength{\abovecaptionskip}{0cm}
\setlength{\belowcaptionskip}{0cm}
\includegraphics[width=\textwidth]{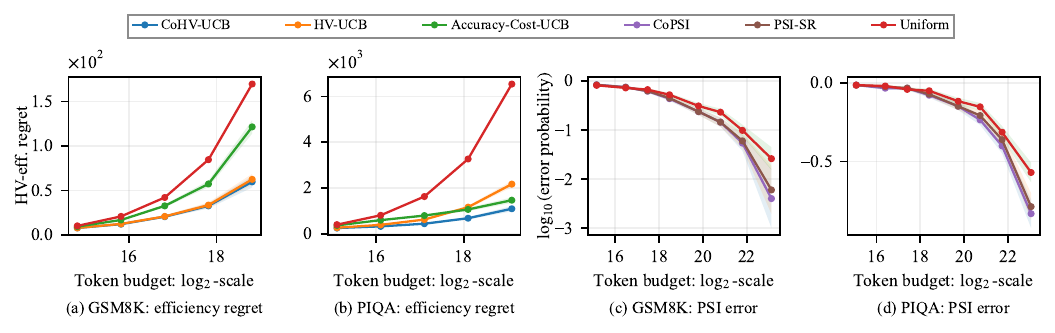}
\caption{Main experimental results. Panels (a)--(b) show hypervolume-efficiency regret over 100 paired replays; panels (c)--(d) show
$\log_{10}\Pr(\widehat{\mathcal P}_B\neq\mathcal P^\star)$ in the structured 14-arm setting over 500 replays. Shaded regions are 95\% confidence intervals.}
\label{fig:main-results}
\end{figure*}

\subsection{Experimental Results}
\paragraph{Online selection.}
The primary measure is hypervolume-efficiency regret, which evaluates the quality obtained per unit cost over the decision process. Standard hypervolume regret, within-run trajectories, and additional diagnostics are provided in Appendix~\ref{app:additional-experiments}.

Figure~\ref{fig:main-results}(a)--(b) shows that adaptive multi-objective selection consistently outperforms uniform allocation as the budget increases. At the largest tested token budgets, \textsc{CoHV-UCB} reduces the hypervolume-efficiency regret from 169.45 to 59.88 on GSM8K and from 6538.22 to 1097.75 on PIQA, corresponding to relative reductions of 64.7\% and 83.2\%, respectively. These results indicate that \textsc{CoHV-UCB} effectively uses the available token budget and avoids repeatedly evaluating configurations with poor multi-objective efficiency.

The improvement of \textsc{CoHV-UCB} over \textsc{HV-UCB} is smaller on GSM8K than on PIQA. This is because rewards and costs are more aligned on GSM8K. Its profiling token costs vary only mildly, from 118.37 to 147.21 tokens, and the configuration with the largest hypervolume also achieves the largest hypervolume-per-cost value. Thus, cost normalization barely changes the arm ranking. In contrast, PIQA has larger cost dispersion, with costs ranging from 95.01 to 152.16 tokens, and its hypervolume-maximizing configuration differs from its efficiency-maximizing configuration. As a result, the cost denominator changes the preferred configuration more substantially, leading to a larger advantage for \textsc{CoHV-UCB}.

\textsc{CoHV-UCB} also consistently improves over \textsc{Accuracy-Cost-UCB}, showing the benefit of optimizing both accuracy and efficiency rather than accuracy alone. Because the horizontal axis displays $\log_2 B$, logarithmic dependence on the budget corresponds to a linear trend in this coordinate. The finite-budget curves may deviate from exact linearity because Theorem~\ref{thm:cohv_ucb_regret} provides an upper bound rather than an equality.

\paragraph{Pareto set identification.}
Figure~\ref{fig:main-results}(c)--(d) reports the base-10 logarithm of the identification error probability in Theorem~\ref{thm:copsi}. The advantage of successive elimination becomes clearer as the budget increases. On GSM8K, \textsc{CoPSI} reduces the error probability from 0.614 to 0.004 over the displayed budget range, compared with 0.662 to 0.026 for uniform allocation. At the largest budget, this corresponds to an 84.6\% relative reduction. On PIQA, \textsc{CoPSI} achieves an error probability of 0.148 at the largest budget, compared with 0.270 for uniform allocation, yielding 45.2\% lower error.

Uniform allocation is competitive at the smallest budgets because it avoids committing to noisy early classifications. With more observations, however, \textsc{CoPSI} can eliminate arms whose status is already clear and concentrate the remaining budget on configurations near the Pareto boundary. This explains why its advantage over uniform allocation widens with the budget. The cost-insensitive PSI-SR method also benefits from successive elimination, but \textsc{CoPSI} attains lower errors at the largest budget, with 0.004 versus 0.006 on GSM8K and 0.148 versus 0.164 on PIQA. The modest difference between the two elimination methods is consistent with the limited cost dispersion in the selected configurations. Additional results and analysis are provided in Appendix~\ref{app:additional-experiments}.

\section{Conclusion and Future Work}
In this paper, we studied LLM configuration evaluation as a cost-aware multi-objective bandit problem under limited evaluation budgets. We considered two complementary objectives: online configuration selection through regret minimization and Pareto-optimal configuration identification under a fixed budget. For online selection, we proposed \textsc{CoHV-UCB}, which selects configurations by optimizing a hypervolume-per-cost index. We proved that its standard budgeted regret bound is $O\left(\sum_{i\ne i^\star}\frac{C_{\lambda,H}^2\log B}{\Delta_i}\right)$, showing logarithmic dependence on the evaluation budget and matching the classical single-objective budgeted bandit result in budget and gap dependence. For fixed-budget identification, we proposed \textsc{CoPSI}, which uses cost-aware sampling and empirical Pareto gaps to identify non-dominated configurations. Its error probability is bounded by $2K^2D\exp\left(-\frac{B}{256L_{K,\lambda}H_{\boldsymbol\mu,c}}\right)$, which recovers the standard fixed-budget Pareto set identification rate when all evaluation costs are equal. Experiments on LLM configuration evaluation tasks further demonstrate the benefit of jointly considering evaluation cost and multiple objectives.

Several directions remain for future work. First, the current formulation treats each LLM configuration as an independent arm; incorporating structural information among models, prompts, and decoding parameters may further improve sample efficiency. Second, extending the framework to contextual or non-stationary settings would be useful for personalized evaluation and evolving deployment environments. Third, more robust Pareto identification algorithms are needed when objective gaps are small or early estimates are unreliable.

\bibliographystyle{plainnat}
\bibliography{ref}

\clearpage
\onecolumn
\appendix
\setcounter{secnumdepth}{1}
\section{Additional Experimental Details and Results}
\label{app:additional-experiments}

\subsection{Experimental Details}

This subsection provides the preprocessing, replay construction, budget
selection, and hyperparameter details used to produce the experimental
results in the main paper and the supplementary material.

We split the 1,000 instances once into 200 profiling and 800 evaluation instances. The profiling split determines the latency percentiles $(\tau_5,\tau_{95})$, the maximum token cost $c_{\max}$, and the deterministic arm cost $c_i$, defined as the profiling mean token count. For configuration $i$ and evaluation instance $q$, a pull returns binary accuracy, token cost $c_{i,q}$, and normalized inverse latency
\[
r^{\rm eff}_{i,q}=\operatorname{clip}\!\left(
\frac{t_{i,q}^{-1}-\tau_{95}^{-1}}
{\tau_{5}^{-1}-\tau_{95}^{-1}},0,1\right).
\]
Online methods internally use $\widetilde c_{i,q}=c_{i,q}/c_{\max}$, while figures report actual token budgets. Feedback is generated by arm-wise bootstrap sampling from the evaluation split, with a shared pre-generated stream for all methods in each repetition.

We parameterize budget by $\rho=B/(K\bar c)$. Online experiments use $\rho\in\{2,4,8,16,32\}$ and 100 paired replays. Their reward and cost confidence radii are
\[
\beta^{r}_{i,t}=s_r\sqrt{\alpha\log(T_B)/n_{i,t}},
\qquad
\beta^{c}_{i,t}=s_c\sqrt{\alpha\log(T_B)/n_{i,t}},
\]
with $n_{\rm init}=\max\{1,\lfloor\eta\rho\rfloor\}$. Profiling-only
selection uses
\[
\begin{aligned}
s_r&\in\{0.003,0.01,0.03,0.1\},&
s_c&\in\{0.001,0.003,0.01\},&
\eta&\in\{0.05,0.1\},
\end{aligned}
\]
and selects $(s_r,s_c,\eta)=(0.01,0.01,0.05)$ on both datasets.
Feasible selection uses the profiling mean cost to avoid exceeding the
remaining budget.

The 14-arm identification subset is formed outcome-independently from GPT-2, CodeLlama-7B, Tulu-7B, Tulu-2-7B, Gemma-7B, LLaMA-7B, and Mistral-7B at temperatures 0.5 and 1.0, with generation length 128 and direct prompting. Its minimum population gaps are 0.005 on GSM8K and 0.010 on PIQA. We use $\rho\in\{20,50,100,200,500,1000,2000,5000\}$ and 500 paired repetitions. Gap ties use relative tolerance $10^{-10}$ and absolute tolerance $10^{-12}$. Complete-set experiments use $\rho\in\{10,20,50,100,200\}$ and 100 repetitions. The stochastic-cost robustness comparison also uses 500 paired repetitions for each method and budget.

\paragraph{Supplementary evaluation metrics.}
For an online trajectory, let
$\Delta_{i_s}=\nu_{i^\star}-\nu_{i_s}$ be the hypervolume-efficiency
gap incurred at pull $s$. The moving-average per-pull regret at pull
$t$, with window length $w$, is
\[
    \overline{\Delta}_{t}^{(w)}
    =
    \frac{1}{\min\{w,t\}}
    \sum_{s=\max\{1,t-w+1\}}^{t}\Delta_{i_s}.
\]
We use $w=50$. Unlike cumulative regret, this metric measures recent
decision quality and is not mechanically increasing with the number of
pulls.

For full-set Pareto recovery, let $\widehat{\mathcal P}_B$ be the
estimated Pareto set and $\mathcal P^\star$ the true set. We report
their F1 score,
\[
    \operatorname{F1}
    =
    \frac{
        2\left|\widehat{\mathcal P}_B\cap\mathcal P^\star\right|
    }{
        \left|\widehat{\mathcal P}_B\right|
        +
        \left|\mathcal P^\star\right|
    },
\]
which is the harmonic mean of Pareto precision and Pareto recall.
It equals one only under exact recovery and provides a graded measure
when ties or very small gaps make exact identification difficult.

\subsection{Additional Experimental Results}

This subsection complements the main results with standard hypervolume
regret, within-run online trajectories, full-set approximate Pareto
recovery, and robustness to stochastic token costs.

\begin{figure*}[!htb]
\centering
\setlength{\abovecaptionskip}{0cm}
\setlength{\belowcaptionskip}{0cm}
\includegraphics[width=\textwidth]{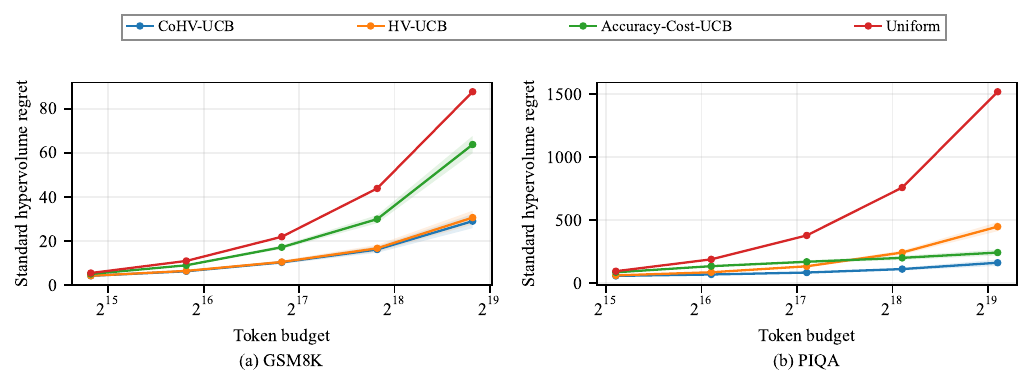}
\caption{Standard hypervolume regret versus actual token budget.}
\label{fig:standard-hv-regret}
\end{figure*}

\paragraph{Standard hypervolume regret.}
Figure~\ref{fig:standard-hv-regret} complements the efficiency-regret
result in the main paper with the standard budgeted metric. Across all
five budgets and both datasets, \textsc{CoHV-UCB} attains the lowest
mean standard regret among the four methods. Its advantage becomes most
pronounced on PIQA as the budget increases. At $\rho=32$, the regrets
of \textsc{CoHV-UCB}, \textsc{HV-UCB},
\textsc{Accuracy-Cost-UCB}, and uniform allocation are 161.07, 447.35,
241.79, and 1517.77, respectively. Thus, \textsc{CoHV-UCB} reduces
standard regret by 64.0\% relative to \textsc{HV-UCB}, by 33.4\%
relative to \textsc{Accuracy-Cost-UCB}, and by 89.4\% relative to
uniform allocation. These comparisons separately demonstrate the value
of accounting for token cost, retaining the efficiency objective, and
adapting the allocation to observed rewards.

The corresponding GSM8K regrets at $\rho=32$ are 29.01, 30.62, 63.75,
and 87.69. The difference between the two hypervolume UCB methods is
smaller because the highest hypervolume arm is also highly competitive
in hypervolume per cost on this dataset. Nevertheless,
\textsc{CoHV-UCB} remains best and reduces regret by 54.5\% relative to
\textsc{Accuracy-Cost-UCB} and by 66.9\% relative to uniform
allocation. The results therefore show that cost aware multi-objective
selection improves not only the efficiency-normalized criterion used
to construct the index, but also the cumulative hypervolume collected
under the same token budget.

\begin{figure*}[!htb]
\centering
\setlength{\abovecaptionskip}{0cm}
\setlength{\belowcaptionskip}{0cm}
\includegraphics[width=\textwidth]{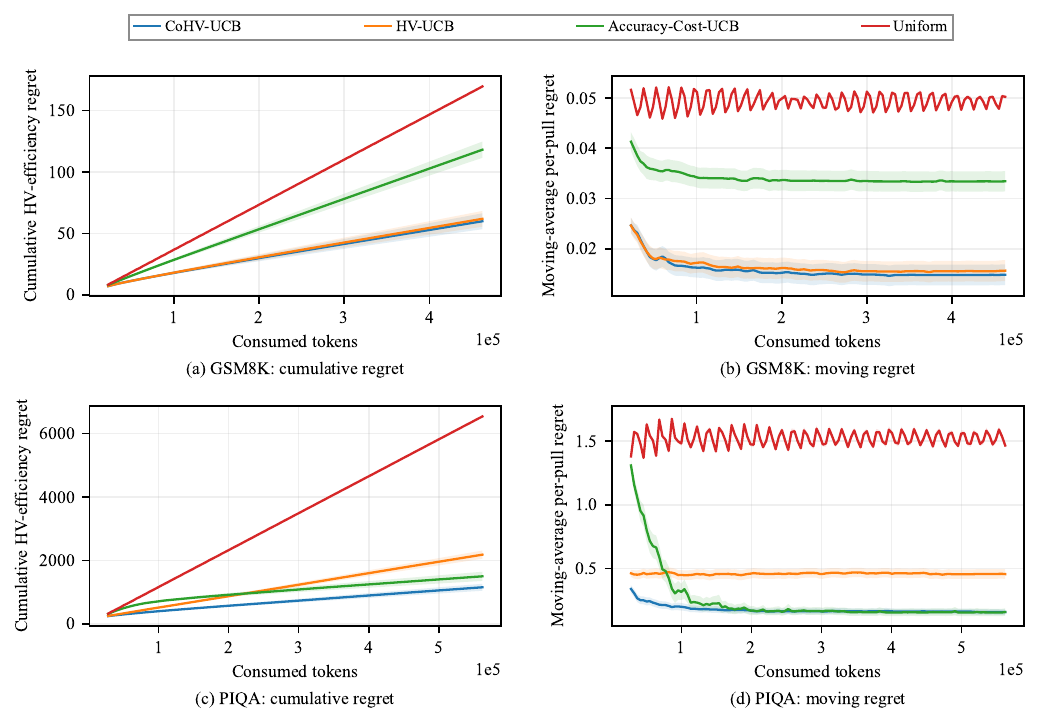}
\caption{Online decision trajectories at the fixed budget $\rho=32$. Shaded regions are 95\% confidence intervals over 100 paired replays. The moving-average panels use the most recent 50 pulls. All horizontal axes report consumed tokens.}
\label{fig:online-trajectory}
\end{figure*}

\paragraph{Online decision trajectories.}
To expose performance during the decision process rather than only at
the stopping budget, Figure~\ref{fig:online-trajectory} fixes
$\rho=32$ and records each method after every pull. Curves are aligned
by consumed tokens, so the methods are compared at equal resource
expenditure even when they complete different numbers of evaluations.
The left column reports cumulative hypervolume-efficiency regret, while
the right column reports the moving average of the per-pull efficiency
gap over the most recent 50 decisions.

On PIQA, the separation emerges early and persists throughout most of
the trajectory. At the end of the displayed budget,
\textsc{CoHV-UCB} has cumulative regret 1149.32, compared with 2179.81
for \textsc{HV-UCB}, 1495.65 for \textsc{Accuracy-Cost-UCB}, and
6544.07 for uniform allocation. Its final moving-average per-pull
regret is 0.156, approximately one third of the 0.455 value of
\textsc{HV-UCB} and about one tenth of the 1.466 value of uniform
allocation. It is also comparable to the 0.157 value of
\textsc{Accuracy-Cost-UCB}, while achieving substantially smaller
cumulative regret. This indicates that \textsc{CoHV-UCB} combines
strong recent decisions with better performance over the full
trajectory.

The GSM8K trajectories exhibit the same ordering against the
single-objective and nonadaptive baselines. The final cumulative regret
of \textsc{CoHV-UCB} is 59.85, versus 118.04 for
\textsc{Accuracy-Cost-UCB} and 169.55 for uniform allocation. The
corresponding moving-average values are 0.0148, 0.0334, and 0.0502.
\textsc{CoHV-UCB} and \textsc{HV-UCB} remain close, with the latter
ending at cumulative regret 61.81 and moving-average regret 0.0156,
which agrees with the limited cost dispersion discussed in the main
paper. Although the moving averages do not monotonically converge to
zero, \textsc{CoHV-UCB} consistently avoids the much larger recent
losses incurred by accuracy-only and uniform selection.

\begin{figure*}[t]
\centering
\setlength{\abovecaptionskip}{0cm}
\setlength{\belowcaptionskip}{0cm}
\includegraphics[width=\textwidth]{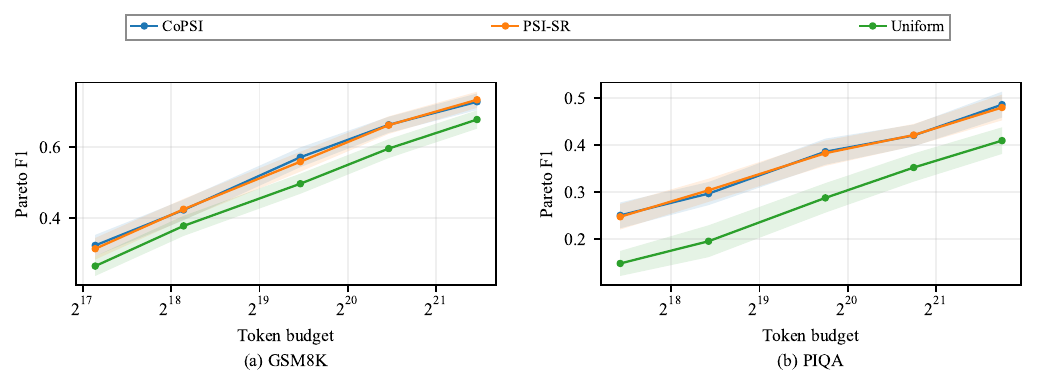}
\caption{Pareto F1 on all 108/132 configurations, where ties make exact identification uninformative.}
\label{fig:full-f1}
\end{figure*}

\paragraph{Approximate recovery on the complete configuration sets.}
The complete GSM8K and PIQA sets contain ties and gaps near zero, making
exact identification uninformative at the available sample size.
Figure~\ref{fig:full-f1} therefore reports Pareto F1 on all 108 and 132
configurations. \textsc{CoPSI} outperforms uniform allocation at every
reported budget on both datasets. The advantage is already visible at
$\rho=10$: the F1 scores are 0.323 versus 0.265 on GSM8K and 0.250
versus 0.148 on PIQA. Hence adaptive elimination is useful even when
the budget is too small for accurate recovery of the full front.

As the budget increases to $\rho=200$, \textsc{CoPSI} reaches 0.727 on
GSM8K and 0.486 on PIQA, whereas uniform allocation reaches only 0.677
and 0.409. These improvements correspond to absolute F1 gains of 0.050
and 0.077, respectively. The larger relative gain on PIQA, 18.7\%,
is particularly notable because PIQA contains more arms and a denser
set of near-boundary configurations. By rejecting configurations whose
status is already clear, \textsc{CoPSI} reserves more evaluations for
the ambiguous part of the front.

\textsc{CoPSI} and the cost-insensitive \textsc{PSI-SR} baseline are
closer, as expected from the moderate cost dispersion in these data.
Their F1 values at $\rho=200$ are 0.727 and 0.732 on GSM8K and 0.486
and 0.480 on PIQA. Across the complete budget range, neither method
uniformly dominates the other, but both consistently improve over
uniform allocation. The important observation is that incorporating
configuration-dependent costs preserves the statistical benefit of
successive elimination while making the allocation compatible with a
token budget.

\paragraph{Robustness to stochastic token costs.}
The theorem-aligned results charge the fixed profiling mean $c_i$ on
every pull. In the robustness setting, all three methods are instead
charged the realized cost $c_{i,q}$. \textsc{CoPSI} continues to use the
profiling estimate $c_i$ in its cost-aware allocation, \textsc{PSI-SR}
retains its cost-insensitive allocation, and uniform allocation adds
only complete sampling rounds so that every arm receives the same
number of observations. All methods use the same pre-generated feedback
stream within each repetition.

\begin{figure*}[h]
\centering
\setlength{\abovecaptionskip}{0cm}
\setlength{\belowcaptionskip}{0cm}
\includegraphics[width=\textwidth]{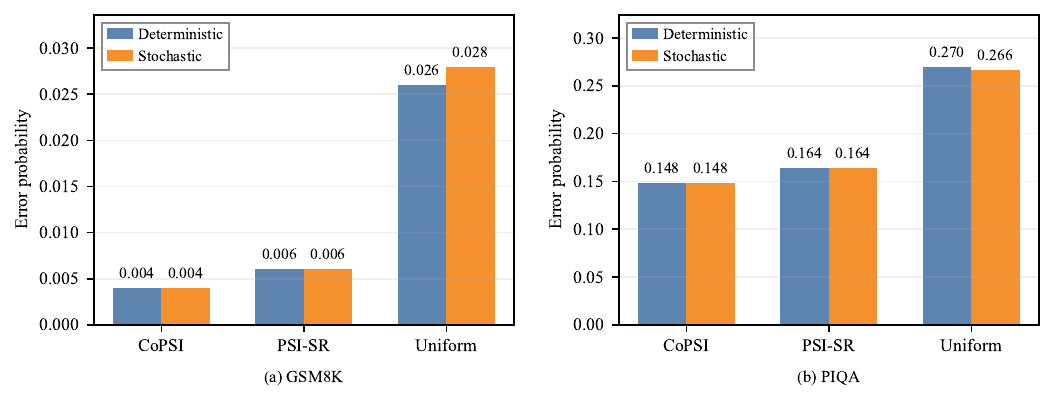}
\caption{Pareto identification error at $\rho=5000$ under deterministic
and stochastic token costs over 500 paired repetitions. Values above
the bars are empirical error probabilities.}
\label{fig:stochastic-cost-robustness}
\end{figure*}

Figure~\ref{fig:stochastic-cost-robustness} shows that realized token
variability causes no material degradation for any method. On GSM8K,
the deterministic and stochastic errors are both 0.004 for
\textsc{CoPSI} and both 0.006 for \textsc{PSI-SR}; uniform allocation
changes only from 0.026 to 0.028. On PIQA, the corresponding values are
0.148 in both settings for \textsc{CoPSI}, 0.164 for \textsc{PSI-SR},
and 0.270 versus 0.266 for uniform allocation. The paired replay
therefore indicates that replacing fixed costs by realized token
charges has negligible influence at this budget. This occurs because
the phase-wise sampling targets of both elimination methods are still
computed from profiling costs. At the large evaluation budget, realized
token fluctuations average out over many pulls and do not activate the
budget stopping rule before these targets are reached. Consequently,
the two cost settings usually evaluate the same feedback samples and
return the same Pareto estimate. Uniform allocation shows a small
difference because its number of complete sampling rounds is determined
directly by the realized costs.

Under stochastic costs, \textsc{CoPSI} retains the lowest error on both
datasets. On GSM8K, its error of 0.004 is 33.3\% below the 0.006 error
of \textsc{PSI-SR} and 85.7\% below the 0.028 error of uniform
allocation. On PIQA, its error of 0.148 is 9.8\% below
\textsc{PSI-SR} and 44.4\% below uniform allocation. These results
support using profiling means for cost-aware allocation when the exact
token charge of an individual evaluation is unknown in advance.

\section{Proof of Theorem~\ref{thm:cohv_ucb_regret}}

\begin{lem}[Optimal benchmark upper bound]
\label{lem:mo_opt_upper}
Consider the budgeted multi-objective bandit problem where pulling arm $i$
incurs a random cost $c_{i,t}\in[\lambda,1]$ with mean $\mu_i^c$ and yields
a scalar hypervolume utility
\[
    H_i
    =
    \operatorname{HV}(\{\bmu_i\}).
\]
Let
\[
    i^\star
    =
    \arg\max_{i\in[K]}
    \frac{H_i}{\mu_i^c},
    \qquad
    \nu_{i^\star}
    =
    \frac{H_{i^\star}}{\mu_{i^\star}^c}.
\]
Assume that the cost samples of each arm are i.i.d.\ across pulls and that,
conditional on the history and the selected arm, the next cost has mean
$\mu_i^c$.  A policy may include the boundary pull that first exhausts the
budget.
Then the optimal expected total hypervolume utility under budget $B$ satisfies
\[
    \mathbb E\!\left[
        \sum_{t=1}^{T(B)} H_{I_t}
    \right]
    \le
    \nu_{i^\star}(B+1).
\]
\end{lem}

\paragraph{Proof.}
Let $I_t$ denote the arm selected
at pull $t$, let $T(B)$ be its number of pulls, and define
\[
    N_i(B)
    =
    \sum_{t=1}^{T(B)}\mathbb I\{I_t=i\}.
\]
By the definition of $i^\star$, every arm satisfies
\[
    H_i
    \le
    \nu_{i^\star}\mu_i^c.
\]
Consequently, the expected utility is bounded as
\[
\begin{aligned}
    \mathbb E\!\left[
        \sum_{t=1}^{T(B)} H_{I_t}
    \right]
    &=
    \sum_{i=1}^{K}H_i\,
        \mathbb E[N_i(B)]
    \\
    &\le
    \nu_{i^\star}
    \sum_{i=1}^{K}\mu_i^c\,
        \mathbb E[N_i(B)].
\end{aligned}
\]
Since $I_t$ is measurable with respect to the history before pull $t$ and
the next cost has conditional mean $\mu_{I_t}^c$, the tower property gives
\[
    \sum_{i=1}^{K}\mu_i^c\,
        \mathbb E[N_i(B)]
    =
    \mathbb E\!\left[
        \sum_{t=1}^{T(B)}c_{I_t,t}
    \right].
\]
Immediately before the last pull, the cumulative cost is at most $B$.
Because every pull costs at most one, including the boundary pull yields
\[
    \sum_{t=1}^{T(B)}c_{I_t,t}
    \le B+1
\]
almost surely.  Thus every admissible policy satisfies
\[
    \mathbb E\!\left[
        \sum_{t=1}^{T(B)} H_{I_t}
    \right]
    \le
    \nu_{i^\star}(B+1).
\]
Taking the supremum over all admissible policies proves the claim.
\hfill$\square$

\paragraph{Proof of Theorem~\ref{thm:cohv_ucb_regret}.}
The proof consists of four steps.

\textbf{Step 1: A uniform concentration event.}
For each arm $i$, objective $d$, and sample size $s$, let $\hat\mu_{i,s}^{(d)}$ denote the empirical mean of the first $s$ observed samples of the $d$-th objective of arm $i$. Similarly, let $\hat\mu_{i,s}^{c}$
denote the empirical mean of the first $s$ cost samples of arm $i$.

Define the good event
\[
    \mathcal E
    =
    \left\{
    \begin{aligned}
    &|\hat\mu_{i,s}^{(d)}-\mu_i^{(d)}|
    \le
    \sqrt{\frac{\alpha\log T_B}{s}},
    \quad
    \forall i\in[K],\ d\in[D],\ s\le T_B,
    \\
    &|\hat\mu_{i,s}^{c}-\mu_i^c|
    \le
    \sqrt{\frac{\alpha\log T_B}{s}},
    \quad
    \forall i\in[K],\ s\le T_B
    \end{aligned}
    \right\}.
\]
Since rewards and costs are bounded in $[0,1]$, Hoeffding's inequality gives, for any fixed $i,d,s$,
\[
    \Pr
    \left(
    |\hat\mu_{i,s}^{(d)}-\mu_i^{(d)}|
    >
    \sqrt{\frac{\alpha\log T_B}{s}}
    \right)
    \le
    2T_B^{-2\alpha}.
\]
The same bound holds for the empirical cost mean. Taking a union bound over all arms, all objectives, the cost coordinate, and all sample sizes $s\le T_B$, we obtain
\[
    \Pr(\mathcal E^c)
    \le
    2K(D+1)T_B\cdot T_B^{-2\alpha}
    =
    2K(D+1)T_B^{1-2\alpha}.
\]
Because each pull costs at least $\lambda$, the total number of pulls before the budget is exhausted is at most $T_B$. Hence the contribution of the failure event to the expected number of pulls of any fixed arm is at most
\[
    T_B\Pr(\mathcal E^c)
    \le
    2K(D+1)T_B^{2-2\alpha}.
\]

\textbf{Step 2: Optimism of the optimal arm.}
On the event $\mathcal E$, for any arm $i$ and any time $t\le T_B$, we have
\[
    \mu_i^{(d)}
    \le
    \bar r_{i,t}^{(d)}+\beta_{i,t},
    \qquad d=1,\ldots,D.
\]
Since $\mu_i^{(d)}\le 1$, the clipping in the definition of $U_{i,t}^{(d)}$ preserves the inequality
\[
    \mu_i^{(d)}
    \le
    U_{i,t}^{(d)}.
\]
Therefore, by the coordinate-wise monotonicity of hypervolume,
\[
    H(\bmu_i)
    \le
    H(\U_{i,t}).
\]
Moreover, on $\mathcal E$,
\[
    \bar c_{i,t}-\beta_{i,t}
    \le
    \mu_i^c.
\]
Since $\lambda/2\le \mu_i^c$, we also have
\[
    \underline c_{i,t}
    =
    \max\left\{\frac{\lambda}{2},\bar c_{i,t}-\beta_{i,t}\right\}
    \le
    \mu_i^c.
\]
Applying these inequalities to the optimal arm $i^\star$ gives
\[
    D_{i^\star,t}
    =
    \frac{H(\U_{i^\star,t})}{\underline c_{i^\star,t}}
    \ge
    \frac{H(\bmu_{i^\star})}{\mu_{i^\star}^c}
    =
    \nu_{i^\star}.
\]
Thus, on the good event, the index of the optimal arm is optimistic.

\textbf{Step 3: Upper bounding the index of a suboptimal arm.}
Fix a suboptimal arm $i\ne i^\star$. We show that once arm $i$ has been pulled sufficiently many times, its index is smaller than the index of the optimal arm on the good event.

On $\mathcal E$, for every coordinate $d$,
\[
    U_{i,t}^{(d)}
    =
    \min\{1,\bar r_{i,t}^{(d)}+\beta_{i,t}\}
    \le
    \mu_i^{(d)}+2\beta_{i,t}.
\]
Therefore,
\[
    \|\U_{i,t}-\bmu_i\|_\infty
    \le
    2\beta_{i,t}.
\]
By the Lipschitz continuity of $H$,
\[
    H(\U_{i,t})-H(\bmu_i)
    \le
    2D\beta_{i,t}.
\]
For the denominator, on $\mathcal E$,
\[
    \bar c_{i,t}\ge \mu_i^c-\beta_{i,t},
\]
and hence
\[
    \bar c_{i,t}-\beta_{i,t}
    \ge
    \mu_i^c-2\beta_{i,t}.
\]
Since
\[
    \underline c_{i,t}
    =
    \max\left\{\frac{\lambda}{2},\bar c_{i,t}-\beta_{i,t}\right\},
\]
we always have
\[
    \underline c_{i,t}\ge \frac{\lambda}{2},
\]
and also
\[
    \underline c_{i,t}
    \ge
    \mu_i^c-2\beta_{i,t}.
\]
The latter inequality implies
\[
    \mu_i^c-\underline c_{i,t}
    \le
    2\beta_{i,t}.
\]

Now consider
\[
    I_{i,t}-\nu_i
    =
    \frac{H(\U_{i,t})}{\underline c_{i,t}}
    -
    \frac{H(\bmu_i)}{\mu_i^c}.
\]
Adding and subtracting
\[
    \frac{H(\bmu_i)}{\underline c_{i,t}},
\]
we obtain
\[
\begin{aligned}
    I_{i,t}-\nu_i
    &=
    \frac{H(\U_{i,t})-H(\bmu_i)}
    {\underline c_{i,t}}
    +
    H(\bmu_i)
    \left(
    \frac{1}{\underline c_{i,t}}
    -
    \frac{1}{\mu_i^c}
    \right)  \\
    &=
    \frac{H(\U_{i,t})-H(\bmu_i)}
    {\underline c_{i,t}}
    +
    \frac{
    H(\bmu_i)(\mu_i^c-\underline c_{i,t})
    }{
    \underline c_{i,t}\mu_i^c
    }.
\end{aligned}
\]
Using
\[
    \underline c_{i,t}\ge \frac{\lambda}{2},
    \qquad
    \mu_i^c\ge \lambda,
\]
together with
\[
    H(\U_{i,t})-H(\bmu_i)
    \le
    2D\beta_{i,t},
\]
\[
    H(\bmu_i)\le 1,
\]
and
\[
    \mu_i^c-\underline c_{i,t}
    \le
    2\beta_{i,t},
\]
we get
\[
\begin{aligned}
    I_{i,t}-\nu_i
    &\le
    \frac{2D\beta_{i,t}}{\lambda/2}
    +
    \frac{2\beta_{i,t}}
    {(\lambda/2)\lambda}  \\
    &=
    \left(
    \frac{4D}{\lambda}
    +
    \frac{4}{\lambda^2}
    \right)
    \beta_{i,t}.
\end{aligned}
\]
Define
\[
    C_{\lambda,H}
    =
    \frac{4D}{\lambda}
    +
    \frac{4}{\lambda^2}.
\]
Then
\[
    I_{i,t}
    \le
    \nu_i
    +
    C_{\lambda,H}\beta_{i,t}.
\]

\textbf{Step 4: Bounding the number of pulls of a suboptimal arm.}
Suppose arm $i\ne i^\star$ is selected at round $t$ on the event $\mathcal E$. Since the algorithm chooses the arm with the largest index,
\[
    I_{i,t}\ge I_{i^\star,t}.
\]
Meanwhile,
\[
    I_{i^\star,t}\ge \nu_{i^\star},\quad 
    I_{i,t}
    \le
    \nu_i+C_{\lambda,H}\beta_{i,t}.
\]
Therefore, if arm $i$ is selected on $\mathcal E$, it must hold that
\[
    \nu_i+C_{\lambda,H}\beta_{i,t}
    \ge
    \nu_{i^\star}.
\]
Equivalently,
\[
    C_{\lambda,H}\beta_{i,t}
    \ge
    \Delta_i.
\]
For convenience, it is enough to require the stronger condition
\[
    C_{\lambda,H}\beta_{i,t}
    \le
    \frac{\Delta_i}{2},
\]
under which arm $i$ cannot be selected.

Since
\[
    \beta_{i,t}
    =
    \sqrt{
    \frac{\alpha\log T_B}{n_{i,t}}
    },
\]
the condition
\[
    C_{\lambda,H}\beta_{i,t}
    \le
    \frac{\Delta_i}{2}
\]
is guaranteed whenever
\[
    n_{i,t}
    \ge
    \frac{4\alpha C_{\lambda,H}^2\log T_B}{\Delta_i^2}.
\]
Therefore, define
\[
    N_i
    =
    \left\lceil
    \frac{4\alpha C_{\lambda,H}^2\log T_B}{\Delta_i^2}
    \right\rceil.
\]
On the event $\mathcal E$, after arm $i$ has been pulled at least $N_i$ times, it cannot be selected again. Thus,
\[
    n_i(B)
    \le
    1+N_i
\]
on $\mathcal E$, where the additional $1$ accounts for the possible boundary pull.

On the complement event $\mathcal E^c$, the number of pulls is at most $T_B$. Therefore,
\[
\begin{aligned}
    \E[n_i(B)]
    &\le
    (1+N_i)\Pr(\mathcal E)
    +
    T_B\Pr(\mathcal E^c) \\
    &\le
    1+N_i
    +
    2K(D+1)T_B^{2-2\alpha}.
\end{aligned}
\]
Substituting the definition of $N_i$ proves the pull-count bound.

Finally, by the definition of hypervolume-efficiency regret,
\[
    R_B^{\mathrm{HV}}
    =
    \sum_{i\ne i^\star}
    \Delta_i\E[n_i(B)].
\]
Using the above pull-count bound gives
\[
    R_B^{\mathrm{HV}}
    =
    O\left(
    \sum_{i\ne i^\star}
    \frac{C_{\lambda,H}^2\log B}{\Delta_i}
    \right),
\]
where we used $T_B=O(B/\lambda)$ and treat $\lambda$ as a problem-dependent constant. This completes the proof.
$\hfill\square$

\section{Proof of Corollary~\ref{cor:standard_cohv_regret}}
\paragraph{Proof.}
The proof follows the standard regret decomposition for budgeted multi-armed bandits, with the scalar reward of arm $i$ replaced by its hypervolume utility $H_i$.

First, by Lemma~\ref{lem:mo_opt_upper} applied with scalar reward mean $H_i$ and cost mean $\mu_i^c$, the optimal expected total hypervolume utility is upper bounded by
\[
    R_B^\star
    \le
    \frac{H_{i^\star}}{\mu_{i^\star}^c}(B+1)
    =
    \nu_{i^\star}(B+1).
\]
Therefore,
\[
\begin{aligned}
    R_B^{\mathrm{std}}
    &=
    R_B^\star
    -
    \E\left[
    \sum_{t=1}^{T(B)}H_{i_t}
    \right]  \\
    &\le
    \nu_{i^\star}(B+1)
    -
    \E\left[
    \sum_{t=1}^{T(B)}H_{i_t}
    \right].
\end{aligned}
\]

Since $H_i$ is deterministic once the arm is fixed, we have
\[
    \E\left[
    \sum_{t=1}^{T(B)}H_{i_t}
    \right]
    =
    \sum_{i=1}^{K}H_i\E[n_i(B)].
\]

Next, denote by
\[
    C_{T(B)}
    =
    \sum_{t=1}^{T(B)}c_{i_t,t}
\]
the total cost consumed before stopping. By the definition of $T(B)$,
\[
    C_{T(B)}\le B
    <
    C_{T(B)+1}.
\]
Since every cost is bounded by $1$, the overshoot is at most one, and thus
\[
    C_{T(B)}>B-1.
\]
Taking expectation and using the standard optional-sampling identity for adaptively sampled bounded costs gives
\[
    \sum_{i=1}^{K}\mu_i^c\E[n_i(B)]
    \ge
    B-1.
\]
Equivalently,
\[
    B+1
    -
    \sum_{i=1}^{K}\mu_i^c\E[n_i(B)]
    \le 2.
\]

Using this inequality, we obtain
\[
\begin{aligned}
    R_B^{\mathrm{std}}
    &\le
    \nu_{i^\star}(B+1)
    -
    \sum_{i=1}^{K}H_i\E[n_i(B)]  \\
    &=
    \nu_{i^\star}
    \left(
    B+1
    -
    \sum_{i=1}^{K}\mu_i^c\E[n_i(B)]
    \right)\\
    &\quad
    +
    \sum_{i=1}^{K}
    \left(
    \nu_{i^\star}\mu_i^c-H_i
    \right)
    \E[n_i(B)] \\
    &\le
    2\nu_{i^\star}
    +
    \sum_{i=1}^{K}
    \left(
    \nu_{i^\star}\mu_i^c-H_i
    \right)
    \E[n_i(B)].
\end{aligned}
\]
For the optimal arm $i^\star$,
\[
    \nu_{i^\star}\mu_{i^\star}^c-H_{i^\star}=0.
\]
For each suboptimal arm $i\ne i^\star$,
\[
    \nu_{i^\star}\mu_i^c-H_i
    =
    \mu_i^c
    \left(
    \nu_{i^\star}-\nu_i
    \right)
    =
    \mu_i^c\Delta_i.
\]
Therefore,
\[
    R_B^{\mathrm{std}}
    \le
    2\nu_{i^\star}
    +
    \sum_{i\ne i^\star}
    \mu_i^c\Delta_i
    \E[n_i(B)].
\]

It remains to substitute the pull-count bound of CoHV-UCB. From the suboptimal-pull analysis, for every $i\ne i^\star$,
\[
    \E[n_i(B)]
    \le
    1
    +
    \left\lceil
      \frac{4\alpha C_{\lambda,H}^2\log T_B}{\Delta_i^2}
    \right\rceil
    +
    2K(D+1)T_B^{2-2\alpha}.
\]
Substituting this inequality into the previous regret decomposition yields the stated finite-budget bound.

Finally, since $\mu_i^c\le 1$, the leading term satisfies
\[
    \mu_i^c\Delta_i
    \cdot
    \frac{C_{\lambda,H}^2\log B}{\Delta_i^2}
    =
    \frac{\mu_i^c C_{\lambda,H}^2\log B}{\Delta_i}
    \le
    \frac{C_{\lambda,H}^2\log B}{\Delta_i}.
\]
This gives the big-$O$ regret bound.
$\hfill\square$

\section{Proof of Theorem~\ref{thm:copsi}}

\paragraph{Proof.}
The proof separates the deterministic classification argument from the
cost-aware sampling calculation. The elimination rule of \textsc{CoPSI}
is the successive-rejects version of Empirical Gap Elimination (EGE):
in each phase, it removes one active arm with the largest empirical
classification gap, and if several arms attain the same largest gap, an
empirically dominated arm is removed whenever possible. The only
difference from standard EGE-SR is the cost-aware choice of the
sampling target $n_r$.

\paragraph{Step 1: Deterministic correctness of the elimination rule.}
We first state the deterministic implication of EGE-SR that will be
used in the proof.

\begin{lem}[Deterministic correctness of EGE-SR]
\label{lem:deterministic-ege-sr}
Consider the successive-rejects version of EGE with the empirical
classification gap defined in \eqref{eq:emp_classification_gap}. At
phase $r$, let $k_r=|\mathcal A_r|$. Suppose that, for every phase $r$
and every active pair $i,j\in\mathcal A_r$, the empirical margins satisfy
\[
    \left|
    \widehat m_r(i,j)-m(i,j)
    \right|
    \le
    \frac{\gamma_{(k_r)}}{8},
    \qquad
    \left|
    \widehat M_r(i,j)-M(i,j)
    \right|
    \le
    \frac{\gamma_{(k_r)}}{8}.
\]
Then every removed arm is classified correctly. Moreover, if one arm
remains active at the end of the procedure, this final surviving arm is
Pareto optimal. Hence the returned set satisfies
\[
    \widehat{\mathcal P}_B=\mathcal P^\star .
\]
\end{lem}

\paragraph{Proof.}
For each dominated arm $i$, fix
\[
    i^\dagger
    \in
    \arg\max_{j\in\mathcal P^\star}m(i,j).
\]
This choice satisfies $m(i,i^\dagger)=\gamma_i$.  Indeed, starting from
any arm that dominates $i$, repeatedly moving to one of its dominators
must terminate at a Pareto arm, and every move can only increase the
coordinate-wise domination margin.  Hence the maximum in
$\Delta_i^-=\max_{j\ne i}m(i,j)$ is attained by a Pareto arm.

Fix a phase $r$, write $k=k_r$, and let
$\varepsilon=\gamma_{(k)}/8$.  We first record three deterministic
consequences of the assumed pairwise error bounds.

\emph{Gap representation.}
For every active arm,
\begin{equation}
\label{eq:emp-gap-max-representation}
    \widehat\gamma_{i,r}
    =
    \max\left\{
        \widehat\Delta_{i,r}^-,
        \widehat\Delta_{i,r}^+
    \right\}.
\end{equation}
If $i$ is empirically dominated, then
$\widehat\Delta_{i,r}^->0$ and
$\widehat\Delta_{i,r}^+\le0$; if it is empirically Pareto optimal, the
two inequalities are reversed.  Thus
\eqref{eq:emp-gap-max-representation} follows directly from
\eqref{eq:emp_classification_gap}.

\emph{Lower bound on active empirical gaps.}
Suppose that every active dominated arm $i$ has its fixed Pareto
dominator $i^\dagger$ active.  Then
\begin{equation}
\label{eq:active-gap-lower-bound}
    \widehat\gamma_{i,r}
    \ge
    \gamma_i-2\varepsilon,
    \qquad i\in\mathcal A_r.
\end{equation}
For a dominated arm, $i^\dagger\in\mathcal A_r$ and the pairwise error
bound give
\[
    \widehat\Delta_{i,r}^-
    \ge
    \widehat m_r(i,i^\dagger)
    \ge
    m(i,i^\dagger)-\varepsilon
    =
    \gamma_i-\varepsilon.
\]
For a Pareto arm $i$, consider any active competitor
$j\in\mathcal A_r\setminus\{i\}$.  The pairwise error bound gives
\[
    \widehat M_r(i,j)
    \ge
    M(i,j)-\varepsilon,
    \qquad
    \widehat M_r(j,i)^+
    \ge
    M(j,i)^+-\varepsilon.
\]
Moreover, the invariant ensures that, whenever $j$ is dominated, an
arm attaining $\Delta_j^-$ remains active; if $j$ is Pareto optimal,
then $(\Delta_j^-)^+=0$.  In either case,
\[
    \left(\widehat\Delta_{j,r}^-\right)^+
    \ge
    \left(\Delta_j^-\right)^+
    -
    \varepsilon.
\]
Consequently,
\[
\begin{aligned}
&\widehat M_r(i,j)
\wedge
\left[
    \widehat M_r(j,i)^+
    +
    \left(\widehat\Delta_{j,r}^-\right)^+
\right]
\\
&\qquad\ge
\left\{
    M(i,j)
    \wedge
    \left[
        M(j,i)^+
        +
        \left(\Delta_j^-\right)^+
    \right]
\right\}
-2\varepsilon.
\end{aligned}
\]
Taking the minimum over active competitors can only increase the
population minimum over all competitors.  Therefore, by the definition
of $\Delta_i^+=\gamma_i$,
\[
    \widehat\Delta_{i,r}^+
    \ge
    \gamma_i-2\varepsilon.
\]
Equation~\eqref{eq:active-gap-lower-bound} follows from
\eqref{eq:emp-gap-max-representation}.

\emph{A missed empirical domination has a small population margin.}
If $i$ is dominated, both $i$ and $i^\dagger$ are active, and
$i^\dagger$ does not empirically dominate $i$, then
\begin{equation}
\label{eq:missed-domination-small-gap}
    \gamma_i
    =
    m(i,i^\dagger)
    \le
    \varepsilon.
\end{equation}
Indeed, the premise implies
$\widehat m_r(i,i^\dagger)\le0$, and the pairwise error bound completes
the claim.

We now establish the dominator-preservation invariant
\begin{equation}
\label{eq:dominator-preservation-invariant}
    \mathcal Q_r:
    \quad
    i\in\mathcal A_r\setminus\mathcal P^\star
    \Longrightarrow
    i^\dagger\in\mathcal A_r
\end{equation}
by induction over the phases.  It holds at $r=1$ because
$\mathcal A_1=[K]$.  Assume that it holds at phase $r$ and, for a
contradiction, suppose that the selected arm is $e_r=i^\dagger$ while
the corresponding dominated arm $i$ remains active.  Among the $k$
active arms there exists an arm $a$ with
$\gamma_a\ge\gamma_{(k)}$.  By
\eqref{eq:active-gap-lower-bound} and the maximality of $e_r$,
\begin{equation}
\label{eq:selected-gap-lower-bound}
    \widehat\gamma_{i^\dagger,r}
    \ge
    \widehat\gamma_{a,r}
    \ge
    \gamma_{(k)}-2\varepsilon
    =
    6\varepsilon.
\end{equation}

If $i^\dagger$ is empirically dominated, its population Pareto
optimality implies $m(i^\dagger,j)\le0$ for every $j$.  Hence
$\widehat\gamma_{i^\dagger,r}
=\widehat\Delta_{i^\dagger,r}^-\le\varepsilon$, contradicting
\eqref{eq:selected-gap-lower-bound}.

Suppose instead that $i^\dagger$ is empirically Pareto optimal.  It
cannot empirically dominate $i$.  Otherwise $i$ is empirically
dominated and, using $i$ as a competitor in the minimum defining
$\widehat\Delta_{i^\dagger,r}^+$,
\[
    \widehat\gamma_{i^\dagger,r}
    \le
    \widehat M_r(i,i^\dagger)^+
    +
    \left(\widehat\Delta_{i,r}^-\right)^+
    =
    \widehat\gamma_{i,r}.
\]
The inequality is either strict, in which case $i^\dagger$ cannot be
selected, or it is an equality, in which case the tie rule selects the
empirically dominated arm $i$.  Both alternatives contradict the
removal of $i^\dagger$ while $i$ remains.

It follows from \eqref{eq:missed-domination-small-gap} that
$\gamma_i\le\varepsilon$.  Moreover,
$M(i,i^\dagger)=-m(i,i^\dagger)=-\gamma_i\le0$, and thus
$\widehat M_r(i,i^\dagger)^+\le\varepsilon$.  The uniform pairwise
bound also gives
\[
    \left(\widehat\Delta_{i,r}^-\right)^+
    \le
    \gamma_i+\varepsilon
    \le
    2\varepsilon.
\]
Using $i$ as a competitor once more yields
\[
    \widehat\gamma_{i^\dagger,r}
    =
    \widehat\Delta_{i^\dagger,r}^+
    \le
    3\varepsilon,
\]
again contradicting \eqref{eq:selected-gap-lower-bound}.  Therefore
$\mathcal Q_{r+1}$ holds, and induction proves
\eqref{eq:dominator-preservation-invariant} at every phase.

It remains to show that the arm removed at each phase is classified
correctly.  The invariant and
\eqref{eq:active-gap-lower-bound} imply, exactly as in
\eqref{eq:selected-gap-lower-bound}, that
\begin{equation}
\label{eq:removed-gap-lower-bound}
    \widehat\gamma_{e_r,r}\ge6\varepsilon.
\end{equation}
If $e_r$ is empirically dominated but truly Pareto optimal, then
$\widehat\gamma_{e_r,r}
=\widehat\Delta_{e_r,r}^-\le\varepsilon$, a contradiction.  If $e_r$
is empirically Pareto optimal but truly dominated, its fixed dominator
$e_r^\dagger$ is active by
\eqref{eq:dominator-preservation-invariant}.  Empirical Pareto
optimality means that $e_r^\dagger$ does not empirically dominate
$e_r$, so \eqref{eq:missed-domination-small-gap} gives
$\gamma_{e_r}\le\varepsilon$.  Since
$M(e_r,e_r^\dagger)=-\gamma_{e_r}\le0$, the pairwise error bound and
the first term in the minimum defining
$\widehat\Delta_{e_r,r}^+$ give
\[
    \widehat\gamma_{e_r,r}
    =
    \widehat\Delta_{e_r,r}^+
    \le
    \widehat M_r(e_r,e_r^\dagger)^+
    \le
    \varepsilon,
\]
again contradicting \eqref{eq:removed-gap-lower-bound}.  Thus every
accepted arm is truly Pareto optimal and every rejected arm is truly
dominated.

After $K-1$ phases, a dominated final survivor would, by
\eqref{eq:dominator-preservation-invariant}, require a distinct active
Pareto dominator.  This is impossible because $\mathcal A_K$ is a
singleton.  The final survivor is therefore Pareto optimal, and the
returned set is exactly $\mathcal P^\star$.
\hfill$\square$

It remains to prove that the cost-aware sampling schedule ensures these
inequalities with high probability.

\paragraph{Step 2: Budget feasibility.}
At phase $r$, \textsc{CoPSI} sets
\[
    n_r
    =
    \left\lfloor
    \frac{B}{L_{K,\lambda}C_r}
    \right\rfloor,
    \qquad
    C_r=C(\mathcal A_r)=\sum_{i\in\mathcal A_r}c_i .
\]
Since exactly one arm is removed in each phase and every arm has
positive cost, $C_r$ is decreasing and therefore $n_r$ is nondecreasing.
The algorithm stops after phase $K-1$, where the last sampled active set
contains two arms; the remaining singleton $\mathcal A_K$ is returned
without further sampling. Thus, the total cost spent by the algorithm is
\[
    \sum_{r=1}^{K-1}
    C_r(n_r-n_{r-1}),
    \qquad n_0=0 .
\]
By summation by parts,
\[
\begin{aligned}
    \sum_{r=1}^{K-1}
    C_r(n_r-n_{r-1})
    &=
    C_{K-1}n_{K-1}
    +
    \sum_{r=1}^{K-2}
    (C_r-C_{r+1})n_r .
\end{aligned}
\]
Using
\[
    n_r
    \le
    \frac{B}{L_{K,\lambda}C_r},
\]
we obtain
\[
\begin{aligned}
    \sum_{r=1}^{K-1}
    C_r(n_r-n_{r-1})
    &\le
    \frac{B}{L_{K,\lambda}}
    +
    \frac{B}{L_{K,\lambda}}
    \sum_{r=1}^{K-2}
    \frac{C_r-C_{r+1}}{C_r}.
\end{aligned}
\]
At phase $r$, the active set has size $k_r=K-r+1$. The removed arm has
cost at most $1$, while the remaining $k_r-1$ active arms have total
cost at least $(k_r-1)\lambda$. Hence
\[
    \frac{C_r-C_{r+1}}{C_r}
    \le
    \frac{1}{1+(k_r-1)\lambda}.
\]
Therefore,
\[
\begin{aligned}
    \sum_{r=1}^{K-1}
    C_r(n_r-n_{r-1})
    &\le
    \frac{B}{L_{K,\lambda}}
    \left(
        1+
        \sum_{r=1}^{K-2}
        \frac{1}{1+(k_r-1)\lambda}
    \right)  \\
    &\le
    \frac{B}{L_{K,\lambda}}
    \left(
        1+
        \sum_{\ell=2}^{K}
        \frac{1}{1+(\ell-1)\lambda}
    \right)
    =
    B .
\end{aligned}
\]
Thus, the sampling schedule is feasible under budget $B$.

\paragraph{Step 3: A phase-wise lower bound on the number of samples.}
By the definition of the cost-aware complexity,
\[
    H_{\boldsymbol\mu,c}
    =
    \max_{\emptyset\ne\mathcal A\subseteq[K]}
    \frac{C(\mathcal A)}{\gamma_{(|\mathcal A|)}^2},
\]
we have, for every realized active set $\mathcal A_r$,
\[
    C_r
    =
    C(\mathcal A_r)
    \le
    H_{\boldsymbol\mu,c}\gamma_{(k_r)}^2 .
\]
Since rewards are bounded in $[0,1]$, every positive classification gap
is at most one. Hence
\[
    C_r\le H_{\boldsymbol\mu,c}.
\]
Under the assumption
\[
    B\ge 2L_{K,\lambda}H_{\boldsymbol\mu,c},
\]
we have
\[
    \frac{B}{L_{K,\lambda}C_r}\ge 2 .
\]
Therefore, using the elementary inequality
$\lfloor x\rfloor\ge x/2$ for all $x\ge2$, we obtain
\[
\begin{aligned}
    n_r
    &=
    \left\lfloor
    \frac{B}{L_{K,\lambda}C_r}
    \right\rfloor
    \ge
    \frac{B}{2L_{K,\lambda}C_r}.
\end{aligned}
\]
Multiplying both sides by $\gamma_{(k_r)}^2$ gives
\begin{equation}
\label{eq:copsi-effective-samples}
\begin{aligned}
    n_r\gamma_{(k_r)}^2
    \ge
    \frac{B\gamma_{(k_r)}^2}
    {2L_{K,\lambda}C_r}
    \ge
    \frac{B}
    {2L_{K,\lambda}H_{\boldsymbol\mu,c}} .
\end{aligned}
\end{equation}

\paragraph{Step 4: Concentration at adaptive sampling targets.}
For each arm $i$ and coordinate $d$, expose in advance an infinite
i.i.d. sequence of rewards with mean $\mu_i^{(d)}$. Let
$\overline\mu_{i,n}^{(d)}$ be the empirical mean of the first $n$
observations in this sequence. Although the active set and the target
$n_r$ are data-dependent, the following time-uniform Hoeffding bound
remains valid:
\begin{equation}
\label{eq:time-uniform-hoeffding-copsi}
    \Pr\left(
    \exists n\ge N:
    \left|
    \overline\mu_{i,n}^{(d)}-\mu_i^{(d)}
    \right|
    >x
    \right)
    \le
    2\exp(-2Nx^2).
\end{equation}
This follows from Ville's inequality applied to the Hoeffding
exponential supermartingales for the upper and lower tails.

For each phase $r$, define the good event
\[
    \mathcal E_r
    =
    \left\{
    \forall i\in\mathcal A_r,\ \forall d\in[D]:
    \left|
    \widehat\mu_{i,r}^{(d)}-\mu_i^{(d)}
    \right|
    \le
    \frac{\gamma_{(k_r)}}{16}
    \right\},
\]
and let
\[
    \mathcal E
    =
    \bigcap_{r=1}^{K-1}\mathcal E_r .
\]
On $\mathcal E_r$, for any active pair $i,j\in\mathcal A_r$, the
empirical pairwise margins satisfy
\[
    \left|
    \widehat m_r(i,j)-m(i,j)
    \right|
    \le
    \frac{\gamma_{(k_r)}}{8},
    \qquad
    \left|
    \widehat M_r(i,j)-M(i,j)
    \right|
    \le
    \frac{\gamma_{(k_r)}}{8},
\]
because each pairwise margin is a minimum or maximum of coordinate-wise
differences and each empirical mean has error at most
$\gamma_{(k_r)}/16$.

We now bound $\Pr(\mathcal E_r^c)$. By
\eqref{eq:copsi-effective-samples}, every realized target $n_r$
satisfies
\[
    n_r
    \ge
    \frac{B}
    {2L_{K,\lambda}H_{\boldsymbol\mu,c}\gamma_{(k_r)}^2}.
\]
Applying \eqref{eq:time-uniform-hoeffding-copsi} with
\[
    x=\frac{\gamma_{(k_r)}}{16},
    \qquad
    N=
    \frac{B}
    {2L_{K,\lambda}H_{\boldsymbol\mu,c}\gamma_{(k_r)}^2},
\]
and taking a union bound over at most $K$ active arms and $D$
coordinates, we obtain
\[
\begin{aligned}
    \Pr(\mathcal E_r^c)
    &\le
    2KD
    \exp\left(
        -2
        \cdot
        \frac{B}
        {2L_{K,\lambda}H_{\boldsymbol\mu,c}\gamma_{(k_r)}^2}
        \cdot
        \frac{\gamma_{(k_r)}^2}{16^2}
    \right)  \\
    &=
    2KD
    \exp\left(
        -
        \frac{B}
        {256L_{K,\lambda}H_{\boldsymbol\mu,c}}
    \right).
\end{aligned}
\]
A union bound over the $K-1$ phases gives
\[
\begin{aligned}
    \Pr(\mathcal E^c)
    &\le
    \sum_{r=1}^{K-1}\Pr(\mathcal E_r^c) \\
    &\le
    2K^2D
    \exp\left(
        -
        \frac{B}
        {256L_{K,\lambda}H_{\boldsymbol\mu,c}}
    \right).
\end{aligned}
\]

\paragraph{Step 5: Concluding the proof.}
On the event $\mathcal E$, the pairwise concentration conditions of
Lemma~\ref{lem:deterministic-ege-sr} hold in every phase. Therefore,
all arms removed by \textsc{CoPSI} are classified correctly, and the
final surviving arm in $\mathcal A_K$ is Pareto optimal. Since
\textsc{CoPSI} returns
\[
    \widehat{\mathcal P}_B
    =
    \widehat{\mathcal P}\cup\mathcal A_K,
\]
we have
\[
    \widehat{\mathcal P}_B=\mathcal P^\star
\]
on $\mathcal E$. Hence
\[
\begin{aligned}
    \Pr\left(
    \widehat{\mathcal P}_B\ne\mathcal P^\star
    \right)
    &\le
    \Pr(\mathcal E^c) \\
    &\le
    2K^2D
    \exp\left(
        -
        \frac{B}
        {256L_{K,\lambda}H_{\boldsymbol\mu,c}}
    \right).
\end{aligned}
\]
This completes the proof.
\hfill$\square$





\end{document}